\documentclass[letterpaper]{article} 
\usepackage{aaai25}  
\usepackage{times}  
\usepackage{helvet}  
\usepackage{courier}  
\usepackage[hyphens]{url}  
\usepackage{graphicx} 
\usepackage{natbib}  
\usepackage{caption} 
\usepackage{algorithm}
\usepackage{algorithmic}

\usepackage{newfloat}
\usepackage{listings}
\DeclareCaptionStyle{ruled}{labelfont=normalfont,labelsep=colon,strut=off} 
\floatstyle{ruled}
\newfloat{listing}{tb}{lst}{}
\floatname{listing}{Listing}
\usepackage{multirow}
\usepackage{booktabs}
\usepackage{amssymb}
\usepackage{makecell}

\usepackage{multibib}
\newcites{supp}{References}

\title{Unlocking the Potential of Reverse Distillation for Anomaly Detection}
\author{
    Xinyue Liu\textsuperscript{\rm 1},
    Jianyuan Wang\textsuperscript{\rm 2}\thanks{Corresponding author.},
    Biao Leng\textsuperscript{\rm 1},
    Shuo Zhang\textsuperscript{\rm 3}
}
\affiliations{
    \textsuperscript{\rm 1}School of Computer Science and Engineering, Beihang University\\
    \textsuperscript{\rm 2}School of Intelligence Science and Technology, University of Science and Technology Beijing\\
    \textsuperscript{\rm 3}Beijing Key Lab of Traffic Data Analysis and Mining, School of Computer \& Technology, Beijing Jiaotong University\\
    \{liuxinyue7, lengbiao\}@buaa.edu.cn, wangjianyuan@ustb.edu.cn, zhangshuo@bjtu.edu.cn
}

\usepackage{bibentry}

\begin{document}
\frenchspacing

\maketitle

\begin{abstract}
Knowledge Distillation (KD) is a promising approach for unsupervised Anomaly Detection (AD). However, the student network's over-generalization often diminishes the crucial representation differences between teacher and student in anomalous regions, leading to  detection failures. To addresses this problem, the widely accepted Reverse Distillation (RD) paradigm designs asymmetry teacher and student network, using an encoder as teacher and a decoder as student. 
Yet, the design of RD 
does not ensure that the teacher encoder effectively distinguishes between normal and abnormal features or that the student decoder generates anomaly-free features. Additionally, the absence of skip connections results in a loss of fine details during feature reconstruction.
To address these issues, we propose RD with Expert, which introduces a novel Expert-Teacher-Student network for simultaneous distillation of both the teacher encoder and student decoder. The added expert network enhances the student's ability to generate normal features and optimizes the teacher's differentiation between normal and abnormal features, reducing missed detections. Additionally, Guided Information Injection is designed to filter and transfer features from teacher to student, improving detail reconstruction and minimizing false positives. Experiments on several benchmarks prove that our method outperforms existing unsupervised AD methods under RD paradigm, fully unlocking RD’s potential.

\end{abstract}

%
\begin{links}
    \link{Code}{https://github.com/hito2448/URD}
\end{links}

\section{Introduction}

Anomaly Detection (AD) is one of the key tasks in
industry.
Due to the difficulty in obtaining anomalous images and the high cost of labeling, unsupervised Anomaly Detection has been extensively studied. Unsupervised AD uses only normal images during training, enabling the model to detect and localize anomalies in the test images. In recent years, thanks to the application of techniques such as reconstruction models, diffusion models, normalizing flow, and knowledge distillation, unsupervised AD has seen rapid advancements.

\begin{figure}[t]
\centering
\includegraphics[width=\columnwidth]{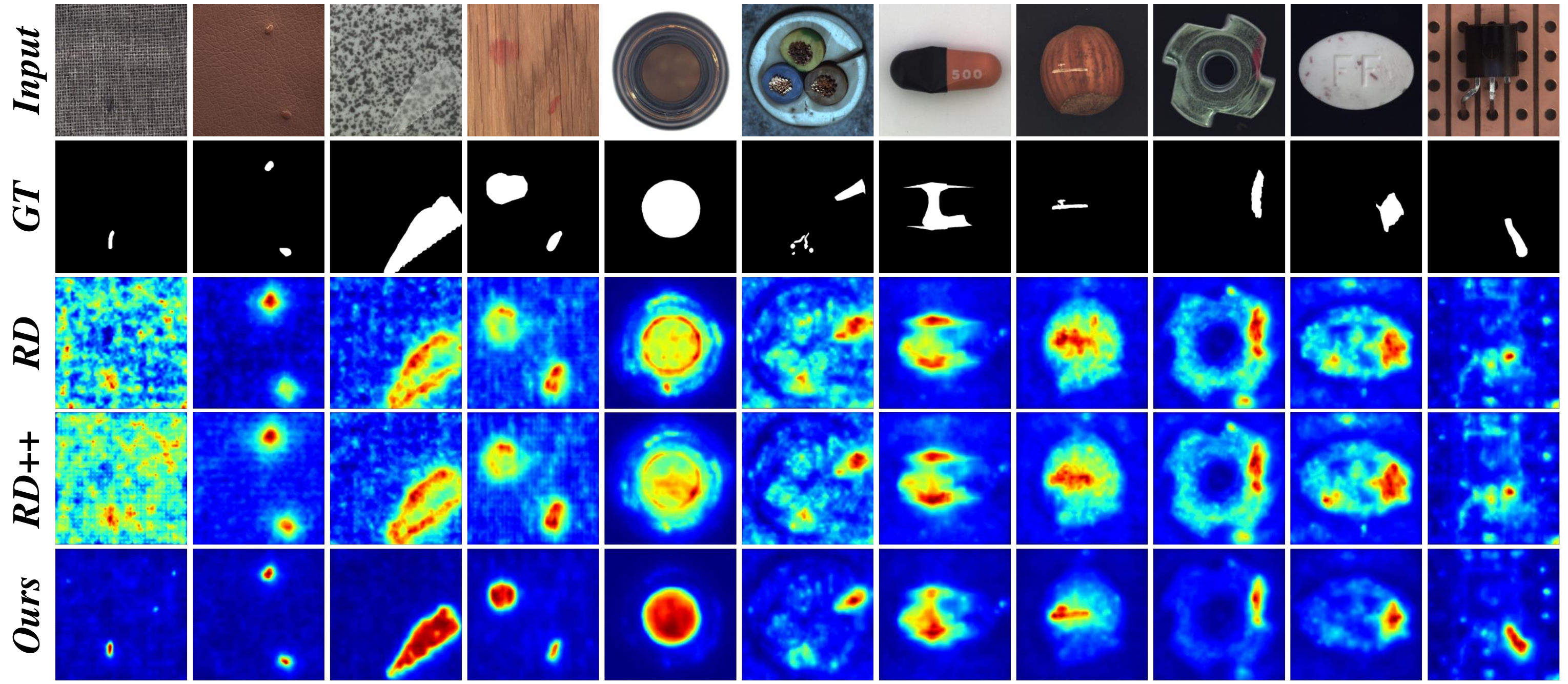} 
\caption{
Anomaly localization examples.
Our method 
reduces missed detections and false positives in RD.}
\label{fig1}
\end{figure}

Knowledge distillation is one of the common paradigms for unsupervised AD \cite{bergmann2020uninformed}. Similar to traditional knowledge distillation, KD-based AD methods typically rely on a teacher-student network, where an initialized student network is distilled by the pre-trained teacher network. Since the student network is trained only on normal images, it is unable to obtain the teacher network's anomaly representation ability. Therefore, during inference, the difference in feature representations between the teacher and the student networks is used to determine whether there are anomalies.
Early KD-based AD methods \cite{bergmann2020uninformed, salehi2021multiresolution, DBLP:conf/bmvc/WangHD021, zhou2022pull} use teacher and student networks with identical or similar architectures and data flow, leading to the student over-generalizing the teacher's anomaly representation ability. To tackle this shortfall, Reverse Distillation (RD) \cite{deng2022anomaly} 
innovatively combines the concept of knowledge distillation with feature reconstruction, using a pre-trained encoder as the teacher and a decoder as the student.
The asymmetric network architectures and the reverse data flow of RD results in better anomaly localization performance.

Although RD is simple and effective for unsupervised AD, there exist some shortcomings in its architectural design: (1) RD's bottleneck module claims to filter out abnormal information so that the student decoder generates anomaly-free features. However, since there is only normal supervision during training, the anomaly filtering is not explicitly guaranteed. Thus, in some cases, the decoder still reconstructs features similar to the teacher encoder's, giving rise to missed detections. (2) To inject encoder features into the decoder for better detail reconstruction from high-level representations, most reconstruction networks introduce skip connection. To prevent anomaly leakage, RD, though as a feature reconstruction network, discards skip connections, limiting its ability to reconstruct fine details and leading to false positives in normal regions.

\begin{figure}[t]
\centering
\includegraphics[width=0.9\columnwidth]{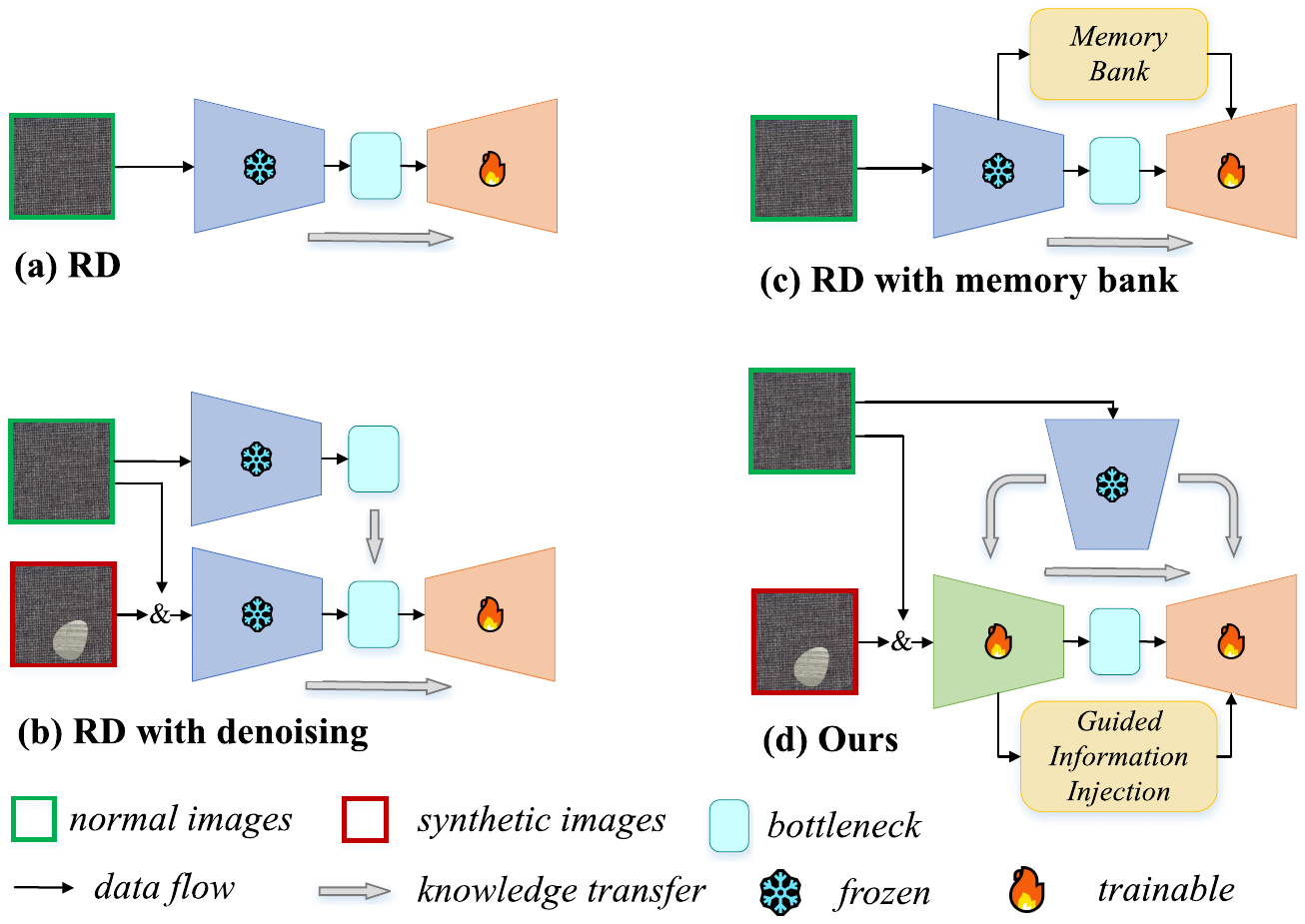} 
\caption{Schematic diagram of the framework and data flow of RD and its variants including our proposed method.}
\label{fig2}
\end{figure}

To address the above issues of RD, recent methods propose improvements from two aspects: \textbf{enriching supervision information} and \textbf{expanding reconstruction information}.
For the lack of anomaly supervision, 
some methods integrates the denoising concept with anomaly synthesis, as shown in in Figure \ref{fig2} (b) \cite{tien2023revisiting, jiang2023masked}. 
For instance, the representative method RD++ \cite{tien2023revisiting} trains the bottleneck to reconstruct the normal feature space from the teacher encoder's abnormal features, thereby ensuring that the student decoder outputs normal features. However, this denoising strategy is based on a strong hypothesis that the features generated by the teacher encoder in anomalous regions differ from the corresponding normal features reconstructed by the student decoder. When the proportion of normal pixels in the receptive field is large, this hypothesis may not hold.
For poor detail reconstruction, other methods  \cite{guo2023template, gu2023remembering} introduces a memory bank to store normal features of the teacher encoder, thus expands the available information of the student decoder during reconstruction, as in Figure \ref{fig2} (c). However, the memory bank brings additional storage requirements, and feature search and alignment also require additional computing requirements.

To unlock the potential of the RD paradigm for unsupervised AD task, we continue to innovate in \textbf{comprehensive supervision} and \textbf{detail reconstruction}. 
First, to ensure the efficacy of 
RD to a great extent, our idea is to incorporate synthetic anomalies to explicitly denoise the student decoder's features while also distill the teacher encoder's features. By enlarging the difference between features generated in normal and anormalous regions, the teacher's anomaly sensitivity is improved.
Additionally, to better reconstruct detail information in lower-level features, 
our intuition is to 
employ similarity attention to directly transfer the teacher's feature information into student, thereby straightforwardly enhancing detail reconstruction.

Unlike the previous RD framework, as illustrated in Figure \ref{fig2} (d), we propose Reverse Distillation with Expert (RD-E) based on an innovative Expert-Teacher-Student Network, which leverages a frozen expert encoder to simultaneously train the teacher encoder and student decoder, enhancing their feature anomaly sensitivity and denoising capability.
In addition, considering that skip connection may cause anomaly leakage, we design Guided Information Injection, utilizing teacher's selective information to aid the student decoder in reconstructing low-level feature details. 
Experimental results on widely benchmarked 
AD datasets demonstrate that anomaly detection and localization performance of our method surpasses that of RD and other mainstream KD-based methods, achieving SOTA.

\section{Related Works}

As one of the crucial tasks in 
industrial quality inspection
\cite{bergmann2019mvtec}, unsupervised Anomaly Detection (AD) has earned increasing attention in recent years. Early methods in unsupervised AD often relies on generative models \cite{bergmann2018improving, akcay2019ganomaly, 
tang2020anomaly, liu2023fair, zhang2023unsupervised}, where the models are trained on normal samples to learn how to reconstruct them and  the reconstruction error is used for inference.
Other methods employs parametric density estimation \cite{defard2021padim, gudovskiy2022cflow, hyun2024reconpatch, zhou2024msflow}, where the parameters of normal distribution are calculated, and anomalies are detected based on how well the samples fit this distribution. Additionally, many methods incorporated pre-trained models \cite{liu2023simplenet, li2023target} and memory banks \cite{roth2022towards, bae2023pni}, comparing the input images to stored normal features. Recently, synthetic anomalies have become a hot topic \cite{li2021cutpaste, lin2024comprehensive}, with external datasets \cite{zavrtanik2021draem} or diffusion models \cite{zhang2024realnet} being used to generate anomalies similar to real-world scenarios, thus aiding unsupervised AD.

Besides, knowledge distillation (KD) based on teacher-student networks has recently been applied to unsupervised AD \cite{bergmann2020uninformed, li2024hyperbolic},
using differences in representation between the two networks 
to identify anomalies. A major concern in KD-based AD is student's over-generalization to teacher's anomaly representations. Some methods 
\cite{salehi2021multiresolution, DBLP:conf/bmvc/WangHD021, rudolph2023asymmetric, liu2024dualmodeling}
address this by using asymmetry teacher and student networks to differentiate their representation abilities. Reverse Distillation (RD) \cite{deng2022anomaly} follows this idea, proposing reverse network architecture and data flow.
Recently, several improvements to RD have been explored \cite{tien2023revisiting, guo2023template, gu2023remembering, jiang2023masked, guo2024recontrast, zhang2024contextual}. 

\begin{figure*}[t]
\centering
\includegraphics[width=0.9\textwidth]{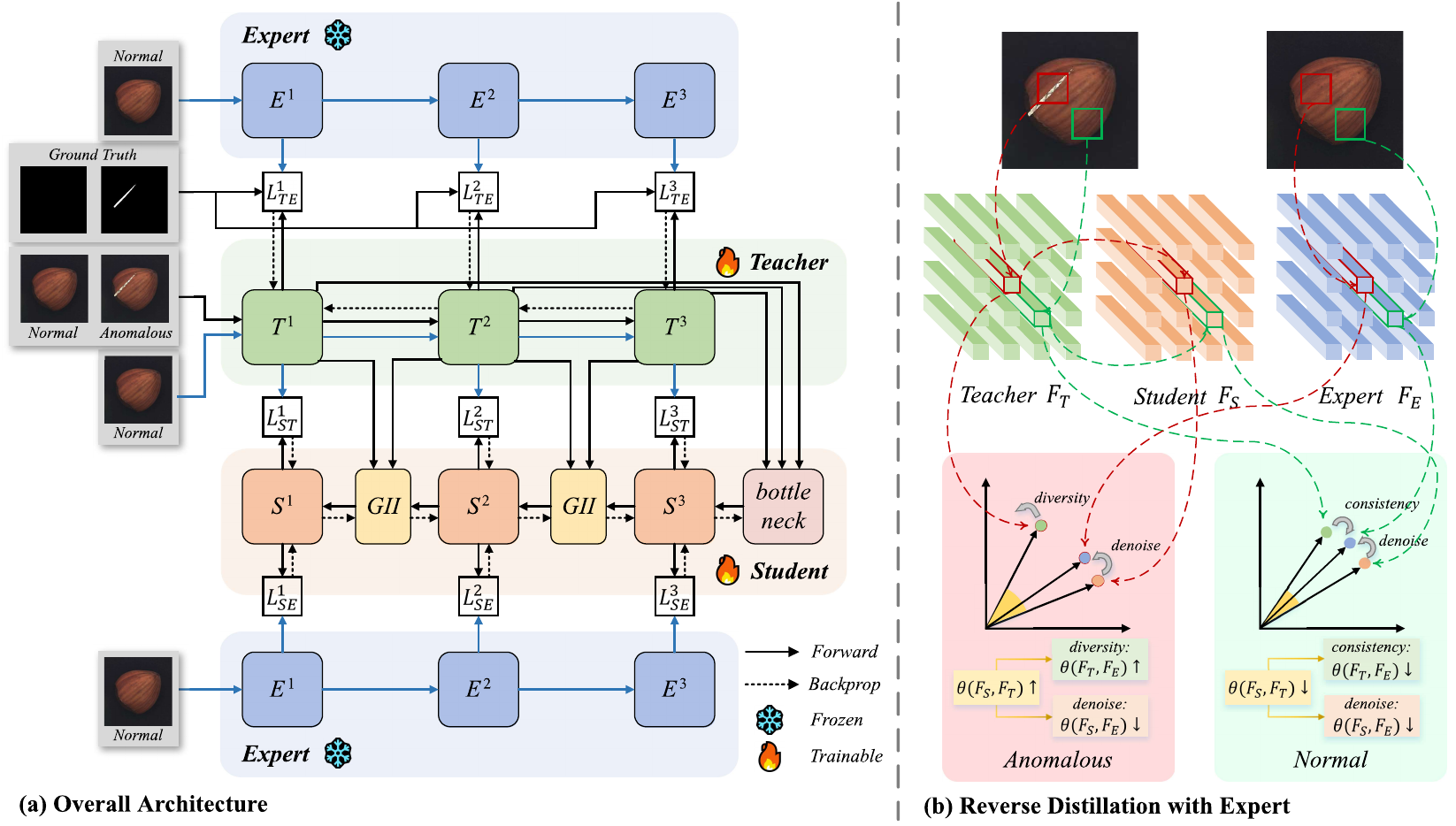} 
\caption{Overview of our proposed method. (a) shows the overall architecture and training process of our designed Expert-Teacher-Student Network, where the expert are frozen and the teacher and student are trainable. Our proposed Guided Information Injection module is inserted between the two blocks of the student. (b) shows how to distill the teacher and student with the expert. The impact of distillation on the teacher and student features is visually represented. Through the two sub-tasks two sub-tasks: making the teacher encoder more sensitive to anomalies and better denoising the student features, differences between teacher and student features are achieved in anomalous regions, while similarities in normal regions are maintained.}
\label{fig3}
\end{figure*}

\section{Revisiting Reverse Distillation}

RD \cite{deng2022anomaly} is a widely adopted unsupervised AD paradigm based on KD. 
The primary components of RD include a pre-trained teacher encoder 
$E$, a one-class bottleneck embedding (OCBE) module,
and a student decoder $D$.

During training, only normal images are used as input. 
The teacher network is frozen, while the bottleneck and student networks are trainable. The OCBE module compresses multi-scale patterns into a low-dimensional space. The compact embedding is then fed into the student network to reconstruct the teacher network’s features. 

During inference, the teacher network can capture abnormal features that deviate from the distribution of normal samples. The OCBE module, by generating compact features, prevents these abnormal perturbations from being input into the student network. Therefore, The student network can generate anomaly-free features regardless of whether normal or abnormal samples are input. The discrepancy in feature reconstruction, measured by their similarity, is used to detect and localize anomalies.

However, the design of RD still has some limitations that affect its anomaly detection performance:

\textbf{(1) Miss Detection Issue}:
The effectiveness of RD relies on two key premises, each with specific requirements for the teacher and student. Firstly, the teacher should be able to capture anomalies by generating abnormal features that differ from normal ones in the anomalous regions. RD assumes this premise is always valid. However, in some cases, such as when the anomalous region is small and normal pixels dominate the receptive field, this assumption may not hold true. Secondly, the student is promised to generate anomaly-free features. Since OCBE essentially performs only a downsampling operation, the generated features used as student input are not guaranteed to be compact and may still contain abnormal information. Additionally, the multi-layer convolutional student decoder has strong generalization capabilities, which means that even if it is trained only on normal samples, it may still generate abnormal features similar to those of the teacher encoder due to over-generalization. 
Consequently, the inability to meet these two key premises results in insufficient difference between the teacher and student features in anomalous regions. Therefore, some anomalies are not be detected.

\textbf{(2) False Positive Issue}:
Since the teacher encoder performs multi-step downsampling and multi-layer convolution, the output high-level features lose lots of details compared with low-level features. Directly using high-level features for low-level feature reconstruction results in reconstruction errors. Most of previous reconstruction networks utilize skip connections to directly pass encoder features to the corresponding decoder layers. However, for RD, this operation may introduce abnormal information from the teacher encoder into the student decoder, making it difficult to generate anomaly-free features. To overcome this challenge, RD designes MFF, which fuses multiple layers of encoder features as the input of the decoder. Although MFF allows low-level features to be included in generating the decoder input, it still downsamples these features to a smaller scale before feature fusion. Hence, some useful detail information is lost, which causes notable discrepancies between student’s reconstructed features and teacher’s features even in normal regions, thereby raising the false positive rate.

\section{Method}

The overall architecture of our proposed method is illustrated in Figure \ref{fig3} (a).
Based on the original teacher-student framework of RD, we design an Expert-Teacher-Student (E-T-S) Network,
which retains the design of the teacher, bottleneck, and student from RD. The teacher encoder $T$ is a WideResNet50 \cite{zagoruyko2016wide} pre-trained on ImageNet \cite{deng2009imagenet}. The bottleneck, named OCBE by RD, includes Multi-scale Feature Fusion (MFF) and One-Class Embedding (OCE) modules. The student decoder $S$ is a symmetric network with $T$, differing in that it replaces downsampling by upsampling. Additionally, $S$ includes Guided Information Injection (GII) to incorporate information from encoder.
Besides, we innovatively introduce an expert network $E$ with the same architecture and initial parameters as $T$.

During training, different from RD, the teacher, bottleneck, and student in E-T-S Network are all trainable. The teacher uses a separate optimizer, while the bottleneck and student share a same optimizer (with the bottleneck being considered part of the student in the following sections).
During inference, the frozen teacher and student are used for anomaly detection and localization.

\subsection{Reverse Distillation with Expert}

The teacher network's ability to perceive anomalies and the student network's capacity to generate anomaly-free features are prerequisites for RD. Previous RD and its variants do not meet both of these two conditions, and trigger missed detection issue.
To tackle this problem, we propose to introduce an expert network to distill both the teacher and the student at the same time, and 
ensure the distillation process covers
\textbf{enhancing the teacher's anomaly sensitivity} and \textbf{denoising the student's features}, as in Figure \ref{fig3} (b).
The teacher network is optimized to 
be more sensitive to anomalies and capable of generating differentiated abnormal and normal features. 
Simultaneously, the student network is trained with a denoising strategy to ensure normal features are generated
even when anomalous samples are input. This dual strategy, based on the introduction of expert network, ensure that the features of teacher and student are similar in normal regions and dissimilar in anomalous regions, which enables effective anomaly detection and localization.

For each normal image $I_n$ in the training set, anomaly synthesis operation is performed to generate a corresponding synthetic anomalous image $I_a$. 
Here, we follow DRÆM \cite{zavrtanik2021draem} for synthesizing anomalies with a Perlin noise generator \cite{perlin} and the Describable Textures Dataset \cite{cimpoi2014describing}.
The teacher $T$ receives a pair of images $I=\{I_n, I_a\}$ as input and outputs three layers of features: $F_T^n = \{{F_T^n}^1,{F_T^n}^2, {F_T^n}^3\} = T(I_n)$ and $F_T^a = \{{F_T^a}^1,{F_T^a}^2, {F_T^a}^3\} = T(I_a)$. The student $S$ takes the features from the teacher network as input and reconstructs the corresponding three features: $F_S^n = \{{F_S^n}^1,{F_S^n}^2, {F_S^n}^3\} = S(F_T^n)$ and $F_S^a = \{{F_S^a}^1,{F_S^a}^2, {F_S^a}^3\} = S(F_T^a)$. The expert $E$, which only takes normal images as input, produces the features that correspond to those of $F_E = \{{F_E^n}^1,{F_E^n}^2, {F_E^n}^3\} = E(I_n)$.

To enhance the teacher's sensitivity to anomalies, we explicitly guide the teacher's feature extraction process using ground truth anomaly masks $M_{gt}$. 
We maintain high cosine similarity for normal regions.
In the meantime, by increasing the cosine distances between the teacher's abnormal features and the expert's normal features in anomalous regions, the teacher network is optimized to generate differentiated abnormal and normal features.
The teacher's training loss $\mathcal{L}_{TE}$ is calculated using L1 distance as
\begin{equation}
\fontsize{9}{10}\selectfont
  {D_{TE}^{n/a}}^i(h,w)=1-\frac{{{F_T^{n/a}}^i(h,w)}^\mathsf{T}\cdot{F_E^{n}}^i(h,w)}{\Vert {F_T^{n/a}}^i(h,w)\Vert\Vert {F_E^n}^i(h,w)\Vert}
\end{equation}
\begin{equation}
\fontsize{9}{10}\selectfont
\mathcal{L}_{TE}^{n/a} =  \sum_{i=1}^{3}\{\frac{1}{H_iW_i}\sum_{h=1}^{H_i}\sum_{w=1}^{W_i}(\vert {D_{TE}^{n/a}}^i(h,w) - M_{gt}^i\vert\}
\end{equation}
\begin{equation}
\fontsize{9}{10}\selectfont
\mathcal{L}_{TE} =  \mathcal{L}_{TE}^n + \mathcal{L}_{TE}^a
\end{equation}
where $H_i$ and $W_i$ represent the height and width of the output feature of the $i$-th encoding block. $M_{gt}^i$ is obtained by downsampling $M_{gt}$ to align the size of $F_T^i$.

To denoise the student's features, we use both the teacher and expert networks to guide the student network, ensuring the student network generates normal features.
To be specific, the student network aims at reconstructing the normal features of the teacher and expert networks whether the input images are normal or anomalous, which is optimized based on cosine similarity with $\mathcal{L}_{S}$ calculated as
\begin{equation}
\fontsize{9}{10}\selectfont
 f = \mathcal{F}(F)
\end{equation}
\begin{equation}
\fontsize{9}{10}\selectfont
 \mathcal{L}_{SE/ST}^i = (1-\frac{{{f_{S}^n}^i}^\mathsf{T} \cdot {f_{E/T}^n}^i}{\Vert {{f_S^n}^i}\Vert\Vert {f_{E/T}^n}^i \Vert})+(1-\frac{{{f_{S}^a}^i}^\mathsf{T} \cdot {f_{E/T}^n}^i}{\Vert {{f_S^a}^i}\Vert\Vert {f_{E/T}^n}^i \Vert})
\end{equation}
\begin{equation}
\fontsize{9}{10}\selectfont
\mathcal{L}_{S} =\sum_{i=1}^{3}(L_{SE}^i + L_{ST}^i)
\end{equation}
where $\mathcal{F}$ is the flatten operation introduced in ReContrast \cite{guo2024recontrast}.

\begin{table*}
\fontsize{9}{10}\selectfont
\centering
\begin{tabular}{c|ccc|ccccc}
    \toprule
 
    & \multicolumn{3}{c|}{Forward Distillation} &  \multicolumn{5}{c}{Reverse Distillation} \\
    &STPM & DeSTSeg &  HypAD&  RD & RD++ & THFR & MemKD&	Ours
\\
    \midrule
             Texture Average& - & 99.1/- & - & 99.7/99.9& \textbf{99.8}/99.9 & 99.7/- & \textbf{99.8}/-  & \textbf{99.8}/\textbf{100} \\
              Object Average   &- & 98.3/-   & - & 98.3/99.3& 98.6/99.4 & 98.9/-  & \textbf{99.5}/- & 98.9/\textbf{99.6} \\
              \midrule
 Total Average  & 95.5/- & 98.6/- & 99.2/99.5& 98.8/99.5& 99.0/99.6 & 99.2/-& \textbf{99.6}/- &  99.2/\textbf{99.7}      \\
  \bottomrule
\end{tabular}
\caption{Image-level anomaly detection results I-AUC/I-AP (\%) on MVTec AD with the best in bold.}
\label{tab:mvtec_detection}
\end{table*}

\subsection{Guided Information Injection} \label{Sec3-3}

\begin{figure}[t]
\centering
\includegraphics[width=\columnwidth]{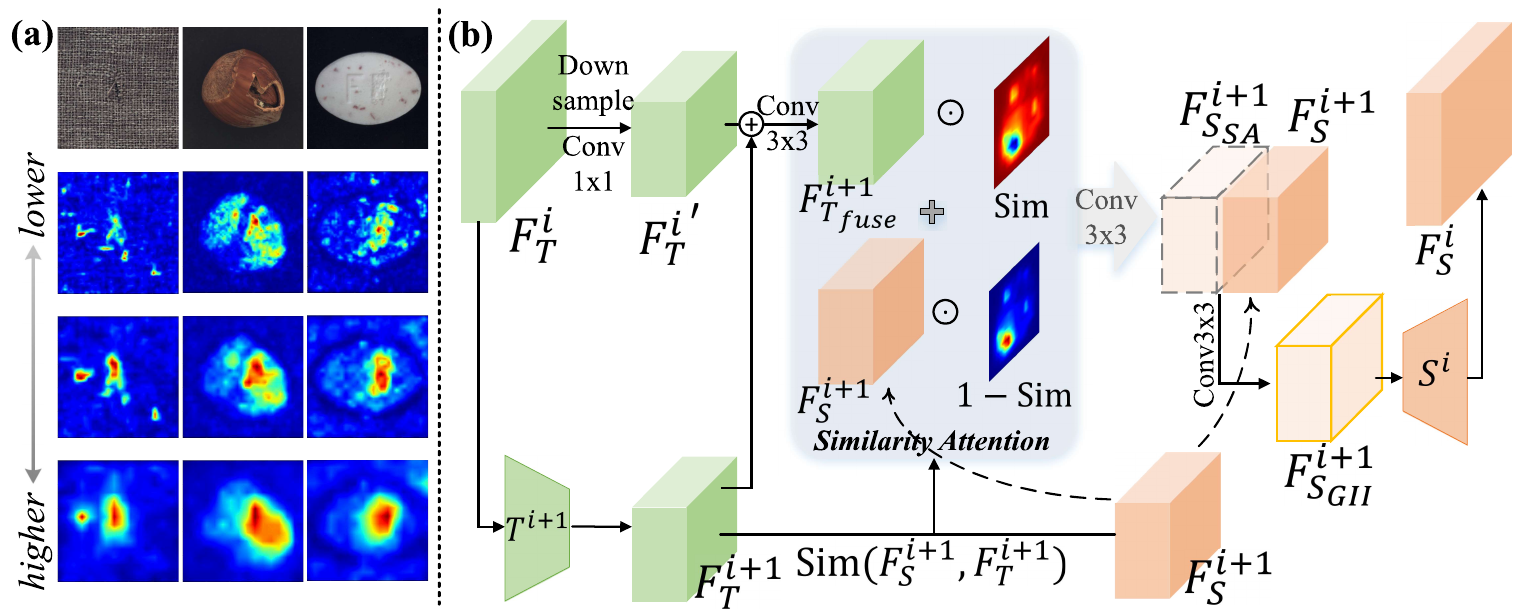} 
\caption{(a) Cosine distance maps between features of teacher and student. (b) Guided Information Injection.}
\label{fig4}
\end{figure}

Considering that: (1) Higher-level features contain less texture details, making detail reconstruction less critical. (2) The shorter generation path for higher-level features naturally leads to better reconstruction quality.
The distance maps calculated by the cosine similarity of higher-level features in Figure \ref{fig4} (a) are therefore believed to effectively locate anomalies and highlight anomalous regions.

Inspired by this, we propose Guided Information Injection (GII), as shown in Figure \ref{fig4} (b).
By leveraging similarity-based attention from higher layers to guide the information injection from encoder to decoder, GII not only directly addresses the issue of the lack of low-level information during reconstruction, but also filters out most of the anomalous information, preventing anomaly leakage.
As a result, detail information is introduced into the decoder in a more controlled and softer manner compared to traditional skip connections.

Specifically, GII is inserted before the student decoder blocks $S^1$ and $S^2$. For the GII module before $S^i$, the input consists of the output features $F_T^{i}$, $F_T^{i+1}$, and $F_S^{i+1}$ from $T^{i}$, $T^{i+1}$, and $S^{i+1}$, respectively. First, $F_T^{i}$ and $F_T^{i+1}$ are adjusted in dimension and combined to obtain the multi-scale fused feature ${F_{T_{fuse}}^{i+1}}$. Then, the cosine similarity between the higher-level features $\mathrm{Sim}(F_T^{i+1}, F_S^{i+1})$ (hereafter referred to as $\mathrm{Sim}$) is calculated, where smaller values indicate a higher likelihood of anomalies. Finally, $\mathrm{Sim}$ is used to control the proportion of the fused feature ${F_T^{i+1}}_{fuse}$ from teacher encoder, and a feature with enriched details ${F_{S_{SA}}^{i+1}}$ is obtained. The original feature $F_S^{i+1}$ and the detail-enriched feature ${F_{S_{SA}}^{i+1}}$ are concatenated and passed to the decoder block $S^{i}$ for subsequent reconstruction, outputting the final feature $F_S^{i}$ with injected detail information. 
For more detailed calculations, see Figure \ref{fig4} (b).

\begin{figure}[t]
\centering
\includegraphics[width=\columnwidth]{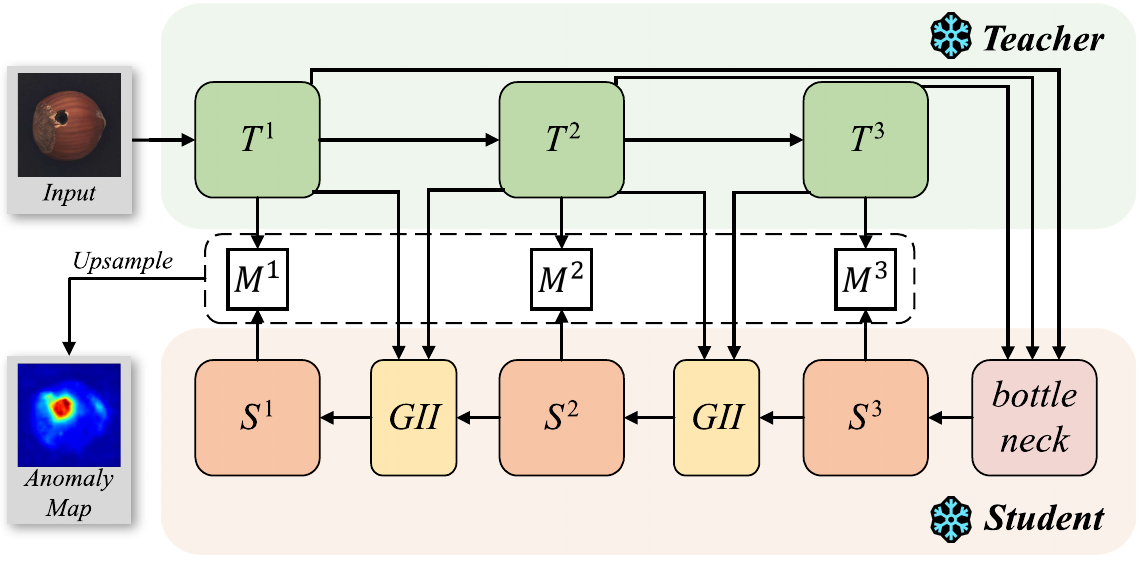} 
\caption{Inference procedure of our proposed method. The expert is removed and both teacher and student are frozen. 
}
\label{fig5}
\end{figure}

\subsection{Inference}

Figure \ref{fig5} illustrates the inference process. During inference, the expert network $E$ is removed, ensuring that our method does not increase storage and computational overhead. 
The approach for anomaly scoring follows RD. For anomaly localization, the score map is obtained by summing the cosine distance maps between the three-layer features of the teacher $T$ and the student $S$ which are upsampled to the input image size. For anomaly detection, the image-level anomaly score is represented by the maximum value in the score map.

\section{Experiments}

\begin{table*}[th]
\setlength{\tabcolsep}{1mm}
\fontsize{9}{11}\selectfont
\centering
  \begin{tabular}{cc|ccc|ccccc}
    \toprule
    \multicolumn{2}{c|}{\multirow{2}{*}{Category}} 
    & \multicolumn{3}{c|}{Forward Distillation} &  \multicolumn{5}{c}{Reverse Distillation} \\
  &  & STPM& DeSTSeg &  HypAD&  RD & RD++ & THFR & MemKD &	Ours
\\
    \midrule
   \multirow{6}{*}{\rotatebox[origin=c]{90}{Textures}}  & Carpet   &  98.8/-/95.8   & 96.1/72.8/93.6  & -/-/92.7 & 99.3/67.2/97.9 & 99.2/63.9/97.7 & 99.2/-/97.7 & 99.1/-/97.5&    \underline{\textbf{99.6}}/\underline{\textbf{83.0}}/\underline{\textbf{98.3}}              \\
              & Grid    & 99.0/-/96.6  & 99.1/\underline{61.5}/96.4 &-/-/\underline{99.7} & 99.3/\textbf{50.2}/\textbf{97.7} & 99.3/49.5/\textbf{97.7} & 99.3/-/\textbf{97.7}  & 99.2/-/96.9 & \underline{\textbf{99.4}}/50.1/97.5                  \\
              & Leather   &  99.3/-/98.0 & \underline{99.7}/\underline{75.6}/99.0 & -/-/99.9 & 99.5/52.6/99.2 & 99.4/51.4/99.2 &  99.4/-/99.2& 99.5/-/99.2 & \underline{\textbf{99.7}}/\textbf{70.0}/\underline{\textbf{99.3}}   \\
              & Tile     &  97.4/-/92.1 &  98.0/90.0/95.5 & -/-/\underline{99.8} &95.8/53.8/91.1 & 96.4/56.2/92.1 & 95.5/-/90.8 & 95.7/-/91.1  &  \underline{\textbf{99.2}}/\underline{\textbf{94.8}}/\textbf{96.8}               \\
              & Wood     &  97.2/-/93.6 & 97.7/\underline{81.9}/\underline{96.1}  & -/-/95.3 &95.3/51.5/93.2 & 95.7/51.8/93.2 & 95.3/-/93.3  & 95.3/-/91.2 & \underline{\textbf{98.1}}/\textbf{80.2}/\textbf{95.2}            \\
              \cmidrule(r){2-10} 
              & Average & 98.3/-/95.2 & 98.1/\underline{76.4}/96.1 & -/-/\underline{97.5} &97.8/55.1/95.8 & 98.0/54.6/96.0 & 97.7/-/95.7   & 97.8/-/95.2 &  \underline{\textbf{99.2}}/\textbf{75.6}/\textbf{97.4} \\
              \midrule
 \multirow{11}{*}{\rotatebox[origin=c]{90}{Objects}} &  Bottle     & 98.8/-/95.1 & 99.2/90.3/96.6 & -/-/\underline{100} & 98.8/78.4/96.9 & 98.7/80.0/96.9 & 98.9/-/97.2 &98.8/-/97.1 & \underline{\textbf{99.3}}/\underline{\textbf{91.6}}/\textbf{97.9}             \\
              & Cable     & 95.5/-/87.7  & 97.3/60.4/86.4 & -/-/93.3 & 97.8/59.6/92.6 & 98.4/63.6/93.9 & 98.5/-/94.8 & 98.3/-/93.4 & \underline{\textbf{98.7}}/\underline{\textbf{73.1}}/\underline{\textbf{94.9}}                \\
              & Capsule   & 98.3/-/92.2 & \underline{99.1}/\underline{56.3}/94.2 & -/-/\underline{96.9} & 98.8/46.6/96.4 & \textbf{98.9}/47.4/96.5 & 98.7/-/95.9 & 98.8/-/96.2 & \textbf{98.9}/\textbf{50.5}/\textbf{96.8}           \\
              & Hazelnut  & 98.5/-/94.3 & \underline{99.6}/\underline{88.4}/97.6 & -/-/\underline{99.7} & \textbf{99.2}/67.9/96.0 & \textbf{99.2}/66.5/\textbf{96.3} & 99.2/-/96.2 & 99.1/-/95.7 & \textbf{99.2}/\textbf{68.0}/96.1            \\
              & Metal\_nut & 97.6/-/94.5 & \underline{98.6}/\underline{93.5}/95.0 & -/-/\underline{98.0} & 97.5/81.8/93.3 & 98.0/\textbf{83.9}/93.2 &  97.4/-/90.5 & 97.2/-/90.8& \textbf{98.4}/83.7/\textbf{93.7}               \\
              & Pill      & 97.8/-/96.5 & \underline{98.7}/83.1/95.3 & -/-/\underline{98.4} & 98.4/80.2/96.9 & 98.4/79.6/97.1 & 98.0/-/96.4  & 98.3/-/96.6 & \underline{\textbf{98.7}}/\underline{\textbf{83.7}}/\textbf{97.5}             \\
              & Screw     & 98.3/-/93.0 & 98.5/\underline{58.7}/92.5 & -/-/95.6 & \underline{\textbf{99.6}}/54.9/\underline{\textbf{98.4}} & \underline{\textbf{99.6}}/\textbf{55.5}/98.3 & 99.5/-/98.2 & \underline{\textbf{99.6}}/-/98.2 & \underline{\textbf{99.6}}/48.8/98.3      \\
              & Toothbrush & 98.9/-/92.2 & \underline{99.3}/\underline{75.2}/94.0 & -/-/\underline{99.9} & 99.1/53.1/94.6 & 99.1/56.3/94.5 & 99.2/-/94.7 & 98.9/-/92.2 &  \underline{\textbf{99.3}}/\textbf{68.5}/\textbf{95.5}        \\
              & Transistor & 82.5/-/69.5 & 89.1/64.8/85.7  & -/-/\underline{100} & 93.1/55.9/79.6 & 94.4/58.3/82.8 &  95.9/-/85.9 & 96.4/-/85.3 & \textbf{\underline{97.5}}/\textbf{\underline{70.3}}/\textbf{90.2}         \\
              & Zipper    & 98.5/-/95.2 & \underline{99.1}/\underline{85.2}/\underline{97.4}  & -/-/94.7 &\textbf{98.9}/61.5/\textbf{96.8} & \textbf{98.9}/60.5/96.4& 98.7/-/96.6 & 98.5/-/95.9 & \textbf{98.9}/\textbf{69.3}/\textbf{96.8}            \\
              \cmidrule(r){2-10} 
              & Average   & 96.5/-/90.9 & 97.9/\underline{75.6}/93.5 & -/-/\underline{97.6} & 98.1/64.0/94.2 & 98.4/65.2/94.6 &  98.4/-/94.6 & 98.4/-/94.1 &  \underline{\textbf{98.9}}/\textbf{70.8}/\textbf{95.8} \\
              \midrule
 \multicolumn{2}{c|}{Total Average} & 97.0/-/92.1 & 97.9/\underline{75.8}/94.4 & 98.0/62.5/\underline{97.6} &98.0/61.0/94.7 & 98.2/61.6/95.1 & 98.2/-/95.0 & 98.2/-/94.5 &  \underline{\textbf{99.0}}/\textbf{72.4}/\textbf{96.3}      \\
  \bottomrule
\end{tabular}
\caption{Pixel-level anomaly localization results P-AUC/P-AP/P-PRO (\%) on MVTec AD with the best KD-based results underlined and the best RD-based results in bold.}
\label{tab:mvtec_loc}
\end{table*}

\begin{table*}
\fontsize{9}{10}\selectfont
\centering
  \begin{tabular}{c|ccccccc}
    \toprule
   Category &  Bracket Black &Bracket Brown & Bracket White &	Connector & Metal Plate & Tubes & Average\\
\midrule
RD & 98.1/6.2/92.1 & 97.2/25.7/95.4 & 99.4/15.6/97.8 & 99.5/64.2/96.9 & 99.1/93.9/96.2 & 99.2/76.0/97.6 & 98.7/46.9/96.0 \\
RD++ & 98.2/9.8/92.8 & 97.1/25.6/94.9  & \textbf{99.5}/12.8/97.2 & 99.3/61.3/96.0 & 99.1/93.3/96.1  &  99.2/74.8/97.4  & 98.7/46.3/95.7\\
MemKD & 97.8/10.7/94.5 & 96.3/20.5/95.2 & 98.8/15.9/97.3 & 99.4/60.6/96.4 & 99.1/94.2/95.2 & 99.2/74.0/97.3 & 98.4/46.1/95.9 \\
Ours & \textbf{98.7}/\textbf{21.4}/\textbf{96.0} & \textbf{98.6}/\textbf{30.3}/\textbf{96.7} & 99.4/\textbf{17.1}/\textbf{98.2} & \textbf{99.6}/\textbf{73.3}/\textbf{97.7} & \textbf{99.2}/\textbf{95.3}/\textbf{96.7} & \textbf{99.4}/\textbf{77.4}/\textbf{98.1} & \textbf{99.2}/\textbf{52.5}/\textbf{97.2} \\
  \bottomrule
\end{tabular}
\caption{Pixel-level anomaly localization results P-AUC/P-AP/P-PRO (\%) on MPDD with the best in bold.}
\label{tab:mpdd_loc}
\end{table*}

\begin{table}
\fontsize{9}{10}\selectfont
\centering
  \begin{tabular}{c|ccc}
    \toprule
   Category &  RD & RD++ &	Ours \\
\midrule
Class 01    & 96.7/50.0/77.7  &  96.1/48.3/71.7 &  \textbf{97.2}/\textbf{55.0}/\textbf{78.6}  \\
Class 02   &  96.8/65.9/66.5  &  96.5/60.1/\textbf{69.4}   &  \textbf{97.4}/\textbf{78.2}/66.9    \\
Class 03  & 99.1/53.5/87.3  &  99.7/59.2/87.2  &  \textbf{99.8}/\textbf{62.5}/\textbf{90.0}    \\
Average  & 97.5/56.5/77.2 &  97.4/55.9/76.1 &   \textbf{98.1}/\textbf{65.2}/\textbf{78.5}        \\
  \bottomrule
\end{tabular}
\caption{Pixel-level anomaly localization results P-AUC/P-AP/P-PRO (\%) on BTAD with the best in bold.}
\label{tab:btad}
\end{table}

\subsection{Experimental Setup}
\subsubsection{Datasets}

We conducted our experiments primarily on \textbf{MVTec AD} \cite{bergmann2019mvtec} containing 5354 images across 15 categories, \textbf{MPDD} \cite{jezek2021deep} containing 1346 images across 6 categories, and \textbf{BTAD} \cite{mishra2021vt}, which includes 2540 images across 3 categories. All datasets have only normal images in the training set, while have both normal and anomalous images in the test set.

\subsubsection{Implementation Details}

A separate detection model is trained for each category. During both training and inference, all images are resized to $256 \times 256$. 
The training batch size is 16, with an early stopping strategy for a maximum of 10,000 iterations. Consistent with RD, the student decoder is optimized using an Adam optimizer with a learning rate of 0.005, while the teacher encoder is trained with another 
one at a learning rate of 0.0001. 
During inference, the anomaly maps are smoothed using a Gaussian filter with $\sigma = 4$.

\subsubsection{Evaluation Metrics}

For anomaly detection, the used evaluation metrics are area under the receiver operating characteristic (AUROC) and average precision (AP). For anomaly localization, in addition to AUROC and AP, we also report per-region-overlap (PRO) \cite{bergmann2020uninformed}.

\subsection{Main Results}

To demonstrate the superiority, we compare our method with various KD-based unsupervised AD methods, including 
STPM \cite{DBLP:conf/bmvc/WangHD021}, 
DeSTSeg \cite{zhang2023destseg}, 
and HypAD \cite{li2024hyperbolic} under Forward Distillation (FD) paradigm, as well as RD \cite{deng2022anomaly}, RD++ \cite{tien2023revisiting}, THFR \cite{guo2023template}, and MemKD \cite{gu2023remembering} under Reverse Distillation  paradigm. For a fairer comparison, we retrain the main comparison methods RD and RD++ under the same environment with our method.

\subsubsection{Anomaly Detection}

Table \ref{tab:mvtec_detection} shows the image-level anomaly detection results on MVTec AD (detailed per-category results are provided in the supplementary materials). For AUC, our method is comparable to the leading KD-based AD methods in overall average.
Regarding AP, our method achieves SOTA performance, with an average of 99.7\% over all categories.

\subsubsection{Anomaly Localization}

We conduct the quantitative comparison of anomaly localization results on MVTec AD in Table \ref{tab:mvtec_loc}. Our method surpasses previous KD-based SOTA in pixel-level AUC, achieving 99.0\%. While our method ranks second in pixel-level AP and PRO metrics with 72.4\% and 96.3\%, it represents the best performance within the RD paradigm. Qualitative visual results are shown in Figure \ref{fig1}.

Furthermore, we extend the quantitative comparison to MPDD and BTAD datasets. Tables \ref{tab:mpdd_loc} and \ref{tab:btad} respectively present the anomaly localization results over all categories on MPDD and BTAD. Our method achieves the best performance in all metrics on the both datasets compared with other RD-based methods, further validating its localization capability.

\subsection{Ablation Analysis}

\begin{table*}
\fontsize{9}{10}\selectfont
\centering
  \begin{tabular}{cc|ccc|ccc|ccc}
    \toprule
    \multirow{2}{*}{Expert} & \multirow{2}{*}{GII} & \multicolumn{3}{c|}{MVTec AD}& \multicolumn{3}{c|}{MPDD}& \multicolumn{3}{c}{BTAD}\\ 
 &    & P-AUC  & P-AP & P-PRO & P-AUC & P-AP & P-PRO & P-AUC & P-AP & P-PRO\\
    \midrule
- & - & 98.02  & 61.01& 94.70 & 98.75 & 46.95 & 95.99 & 97.53& 55.45& 77.17\\
\checkmark & -  &98.69 & 71.80 &96.11 & 98.76& 45.75& 96.06 & 97.97 & 63.63& 77.78\\
- &\checkmark & 98.17 & 60.20 & 94.77 & 99.14& 48.95 &97.17 & 97.83 & 59.14 & 78.23 \\
    \checkmark &\checkmark & 98.97 & 72.37 & 96.31 & 99.14 & 52.46 & 97.25& 98.11 & 65.24 & 78.48 \\
             
  \bottomrule
\end{tabular}
\caption{Ablation localization results (\%) of network composition on MVTec AD, MPDD, and BTAD.}
\label{tab:ablation1}
\end{table*}

\begin{figure*}[t]
\centering
\includegraphics[width=\linewidth]{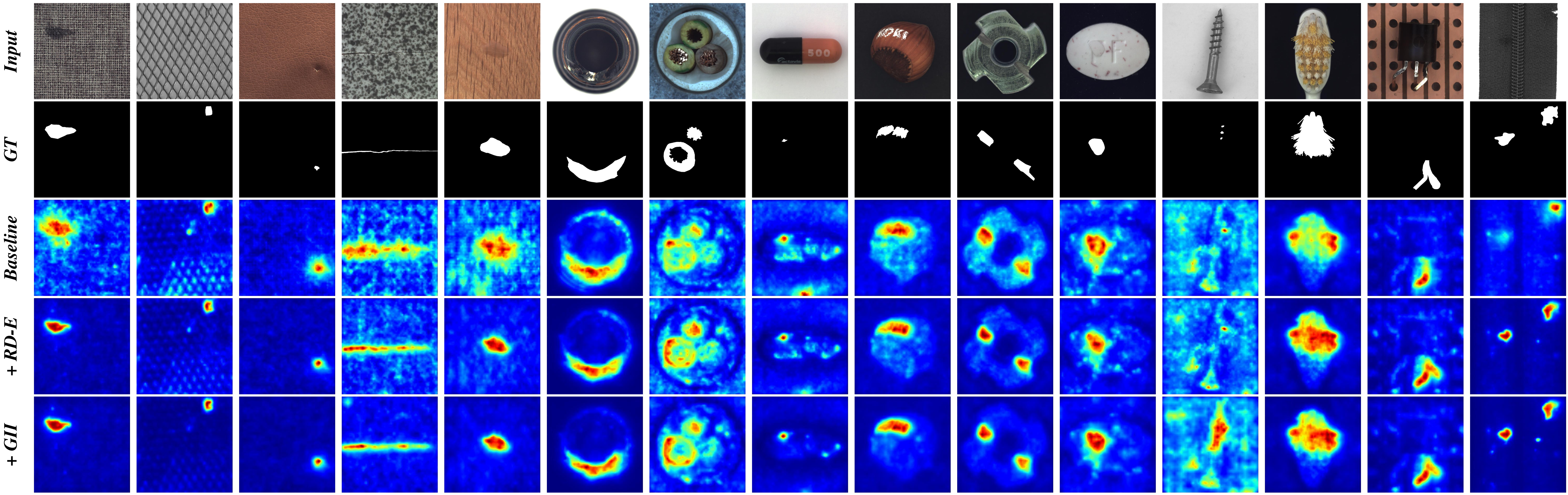} 
\caption{Visualization of ablation study on network composition. 
From top to bottom: the input image, the ground truth masks, the output anomaly maps of Baseline (RD), Baseline+RD-E, and Baseline+RD-E+GII (Ours).
}
\label{fig_vis}
\end{figure*}

\subsubsection{Ablation Study on Network Composition}

Our proposed method primarily includes two innovative components: the design of 
Reverse Distillation with Expert for
distillation supervision innovation and the design of Guided Information Injection for network detail optimization. To demonstrate the effectiveness and necessity of the components, we conduct ablation experiments on MVTec AD, MPDD, and BTAD datasets, as shown in Table \ref{tab:ablation1}. The quantitative results indicate that when both RD-E and GII are applied simultaneously, the method achieves the best  localization results.

In addition, Figure \ref{fig_vis} illustrates the qualitative comparison results, where the baseline refers to the standard RD. It is evident that incorporating RD-E significantly enhances anomaly localization capabilities, reducing the missed detection rate. Furthermore, with the introduction of GII, background noise in anomaly maps in obtained anomaly maps is greatly reduced, leading to a lower false positive rate. These findings align well with our previous analysis.

\begin{table}
\fontsize{9}{10}\selectfont
\centering
  \begin{tabular}{cc|ccccc}
    \toprule
 Den &  Sen & I-AUC & P-AUC & I-AP & P-AP & P-PRO \\
    \midrule
   \multicolumn{7}{c}{Teacher-Student Network} 
\\
\midrule
- & - & 98.77& 98.02 &99.52 & 61.01& 94.70 \\
\checkmark & - &98.62 & 98.29&99.47 & 61.56 & 95.39 \\
    \midrule
 \multicolumn{7}{c}{Expert-Teacher-Student Network} 
\\
\midrule
    \checkmark &\checkmark& 98.72 &98.69 & 99.48& 71.80 &96.11 \\
             
  \bottomrule
\end{tabular}
\caption{Ablation study results (\%) of RD-E on MVTec AD. (Den: Denoising. Sen: Sensitivity.)}
\label{tab:ablation2}
\end{table}

\subsubsection{Ablation Study on Reverse Distillation with Expert
}

In Table \ref{tab:ablation2}, we compare the results on MVTec AD between using only the teacher encoder for the student decoder's feature \textbf{den}oising (Den) and introducing an expert network that enhances the teacher's anomaly \textbf{sen}sitivity while also \textbf{den}oising the student's features (Sen+Den). The results show a significant improvement in anomaly localization when the expert network is added to aid the distillation.

\subsubsection{Ablation Study on Guided Information Injection}

Table \ref{tab:ablation3} presents the results of ablation experiments related to the GII module on MVTec AD. The "+SC" row indicates the absence of similarity attention, where $F_{S_{SA}}^{i+1}=F_{T_{fuse}}^{i+1}$. The "+SA" row shows the results when similarity attention is introduced to filter features. 
The overall results highlight the effectiveness of GII and underscores the importance of the similarity attention mechanism within it.

\begin{table}
\fontsize{9}{10}\selectfont
\centering
  \begin{tabular}{p{5.5mm}c|ccccc}
    \toprule
  && I-AUC & P-AUC & I-AP & P-AP & P-PRO  \\
    \midrule
\multicolumn{2}{c|}{w/o GII} & 98.72 &98.69 & 99.48& 71.80 &96.11  \\
 \midrule
\multirow{2}{*}{w/ GII}& + SC &98.96 & 98.86& 99.56& 71.14 & 96.19 \\
&+ SA & 99.22 & 98.97 & 99.74 & 72.37 & 96.31 \\
  \bottomrule
\end{tabular}
\caption{Ablation study results (\%) of 
GII
on MVTec AD. (SC: Naive skip connection. SA: Similarity attention.)}
\label{tab:ablation3}
\end{table}

\section{Conclusion}

In this paper, we first improves Reverse Distillation with Expert for unsupervised AD. Building on the RD paradigm, we introduce an expert network that distills both the teacher and student networks, ensuring the effectiveness of RD by enhancing the teacher's sensitivity to anomalies and maintaining the student's ability to produce normal features. Besides, to address the challenge of detail reconstruction, we design Guided Information Injection, which uses high-level feature similarity as attention to guide the injection of teacher's features into the student. With these innovations, our method effectively reduces missed detections and false positives in RD, as confirmed by experimental results.


\bibliography{ref}

\begin{thebibliography}{39}
\providecommand{\natexlab}[1]{#1}

\bibitem[{Akcay, Atapour-Abarghouei, and Breckon(2019)}]{akcay2019ganomaly}
Akcay, S.; Atapour-Abarghouei, A.; and Breckon, T.~P. 2019.
\newblock Ganomaly: Semi-supervised anomaly detection via adversarial training.
\newblock In \emph{Computer Vision--ACCV 2018: 14th Asian Conference on Computer Vision, Perth, Australia, December 2--6, 2018, Revised Selected Papers, Part III 14}, 622--637. Springer.

\bibitem[{Bae, Lee, and Kim(2023)}]{bae2023pni}
Bae, J.; Lee, J.-H.; and Kim, S. 2023.
\newblock Pni: industrial anomaly detection using position and neighborhood information.
\newblock In \emph{Proceedings of the IEEE/CVF International Conference on Computer Vision}, 6373--6383.

\bibitem[{Bergmann et~al.(2019)Bergmann, Fauser, Sattlegger, and Steger}]{bergmann2019mvtec}
Bergmann, P.; Fauser, M.; Sattlegger, D.; and Steger, C. 2019.
\newblock MVTec AD--A comprehensive real-world dataset for unsupervised anomaly detection.
\newblock In \emph{Proceedings of the IEEE/CVF conference on computer vision and pattern recognition}, 9592--9600.

\bibitem[{Bergmann et~al.(2020)Bergmann, Fauser, Sattlegger, and Steger}]{bergmann2020uninformed}
Bergmann, P.; Fauser, M.; Sattlegger, D.; and Steger, C. 2020.
\newblock Uninformed students: Student-teacher anomaly detection with discriminative latent embeddings.
\newblock In \emph{Proceedings of the IEEE/CVF conference on computer vision and pattern recognition}, 4183--4192.

\bibitem[{Bergmann et~al.(2018)Bergmann, L{\"o}we, Fauser, Sattlegger, and Steger}]{bergmann2018improving}
Bergmann, P.; L{\"o}we, S.; Fauser, M.; Sattlegger, D.; and Steger, C. 2018.
\newblock Improving unsupervised defect segmentation by applying structural similarity to autoencoders.
\newblock \emph{arXiv preprint arXiv:1807.02011}.

\bibitem[{Cimpoi et~al.(2014)Cimpoi, Maji, Kokkinos, Mohamed, and Vedaldi}]{cimpoi2014describing}
Cimpoi, M.; Maji, S.; Kokkinos, I.; Mohamed, S.; and Vedaldi, A. 2014.
\newblock Describing textures in the wild.
\newblock In \emph{Proceedings of the IEEE conference on computer vision and pattern recognition}, 3606--3613.

\bibitem[{Defard et~al.(2021)Defard, Setkov, Loesch, and Audigier}]{defard2021padim}
Defard, T.; Setkov, A.; Loesch, A.; and Audigier, R. 2021.
\newblock Padim: a patch distribution modeling framework for anomaly detection and localization.
\newblock In \emph{International Conference on Pattern Recognition}, 475--489. Springer.

\bibitem[{Deng and Li(2022)}]{deng2022anomaly}
Deng, H.; and Li, X. 2022.
\newblock Anomaly detection via reverse distillation from one-class embedding.
\newblock In \emph{Proceedings of the IEEE/CVF Conference on Computer Vision and Pattern Recognition}, 9737--9746.

\bibitem[{Deng et~al.(2009)Deng, Dong, Socher, Li, Li, and Fei-Fei}]{deng2009imagenet}
Deng, J.; Dong, W.; Socher, R.; Li, L.-J.; Li, K.; and Fei-Fei, L. 2009.
\newblock Imagenet: A large-scale hierarchical image database.
\newblock In \emph{2009 IEEE conference on computer vision and pattern recognition}, 248--255. Ieee.

\bibitem[{Gu et~al.(2023)Gu, Liu, Chen, Yi, Zhang, Wang, Wang, Shu, Jiang, and Ma}]{gu2023remembering}
Gu, Z.; Liu, L.; Chen, X.; Yi, R.; Zhang, J.; Wang, Y.; Wang, C.; Shu, A.; Jiang, G.; and Ma, L. 2023.
\newblock Remembering Normality: Memory-guided Knowledge Distillation for Unsupervised Anomaly Detection.
\newblock In \emph{Proceedings of the IEEE/CVF International Conference on Computer Vision}, 16401--16409.

\bibitem[{Gudovskiy, Ishizaka, and Kozuka(2022)}]{gudovskiy2022cflow}
Gudovskiy, D.; Ishizaka, S.; and Kozuka, K. 2022.
\newblock Cflow-ad: Real-time unsupervised anomaly detection with localization via conditional normalizing flows.
\newblock In \emph{Proceedings of the IEEE/CVF winter conference on applications of computer vision}, 98--107.

\bibitem[{Guo et~al.(2023)Guo, Ren, Fu, Wang, Zhang, Lan, Wang, and Hou}]{guo2023template}
Guo, H.; Ren, L.; Fu, J.; Wang, Y.; Zhang, Z.; Lan, C.; Wang, H.; and Hou, X. 2023.
\newblock Template-guided Hierarchical Feature Restoration for Anomaly Detection.
\newblock In \emph{Proceedings of the IEEE/CVF International Conference on Computer Vision}, 6447--6458.

\bibitem[{Guo et~al.(2024)Guo, Jia, Zhang, Li et~al.}]{guo2024recontrast}
Guo, J.; Jia, L.; Zhang, W.; Li, H.; et~al. 2024.
\newblock Recontrast: Domain-specific anomaly detection via contrastive reconstruction.
\newblock \emph{Advances in Neural Information Processing Systems}, 36.

\bibitem[{Hyun et~al.(2024)Hyun, Kim, Jeon, Kim, Bae, and Kang}]{hyun2024reconpatch}
Hyun, J.; Kim, S.; Jeon, G.; Kim, S.~H.; Bae, K.; and Kang, B.~J. 2024.
\newblock ReConPatch: Contrastive patch representation learning for industrial anomaly detection.
\newblock In \emph{Proceedings of the IEEE/CVF Winter Conference on Applications of Computer Vision}, 2052--2061.

\bibitem[{Jezek et~al.(2021)Jezek, Jonak, Burget, Dvorak, and Skotak}]{jezek2021deep}
Jezek, S.; Jonak, M.; Burget, R.; Dvorak, P.; and Skotak, M. 2021.
\newblock Deep learning-based defect detection of metal parts: evaluating current methods in complex conditions.
\newblock In \emph{2021 13th International congress on ultra modern telecommunications and control systems and workshops (ICUMT)}, 66--71. IEEE.

\bibitem[{Jiang, Cao, and Shen(2023)}]{jiang2023masked}
Jiang, Y.; Cao, Y.; and Shen, W. 2023.
\newblock A masked reverse knowledge distillation method incorporating global and local information for image anomaly detection.
\newblock \emph{Knowledge-Based Systems}, 280: 110982.

\bibitem[{Li et~al.(2021)Li, Sohn, Yoon, and Pfister}]{li2021cutpaste}
Li, C.-L.; Sohn, K.; Yoon, J.; and Pfister, T. 2021.
\newblock Cutpaste: Self-supervised learning for anomaly detection and localization.
\newblock In \emph{Proceedings of the IEEE/CVF conference on computer vision and pattern recognition}, 9664--9674.

\bibitem[{Li et~al.(2024)Li, Chen, Xu, and Hu}]{li2024hyperbolic}
Li, H.; Chen, Z.; Xu, Y.; and Hu, J. 2024.
\newblock Hyperbolic Anomaly Detection.
\newblock In \emph{Proceedings of the IEEE/CVF Conference on Computer Vision and Pattern Recognition}, 17511--17520.

\bibitem[{Li et~al.(2023)Li, Hu, Li, Chen, Zheng, and Shen}]{li2023target}
Li, H.; Hu, J.; Li, B.; Chen, H.; Zheng, Y.; and Shen, C. 2023.
\newblock Target before shooting: Accurate anomaly detection and localization under one millisecond via cascade patch retrieval.
\newblock \emph{arXiv preprint arXiv:2308.06748}.

\bibitem[{Lin and Yan(2024)}]{lin2024comprehensive}
Lin, J.; and Yan, Y. 2024.
\newblock A Comprehensive Augmentation Framework for Anomaly Detection.
\newblock In \emph{Proceedings of the AAAI Conference on Artificial Intelligence}, volume~38, 8742--8749.

\bibitem[{Liu et~al.(2023{\natexlab{a}})Liu, Li, Du, Jiang, Geng, Wang, and Zhao}]{liu2023fair}
Liu, T.; Li, B.; Du, X.; Jiang, B.; Geng, L.; Wang, F.; and Zhao, Z. 2023{\natexlab{a}}.
\newblock Fair: frequency-aware image restoration for industrial visual anomaly detection.
\newblock \emph{arXiv preprint arXiv:2309.07068}.

\bibitem[{Liu et~al.(2024)Liu, Wang, Leng, and Zhang}]{liu2024dualmodeling}
Liu, X.; Wang, J.; Leng, B.; and Zhang, S. 2024.
\newblock Dual-modeling decouple distillation for unsupervised anomaly detection.
\newblock In \emph{Proceedings of the 32nd ACM International Conference on Multimedia}, 5035--5044.

\bibitem[{Liu et~al.(2023{\natexlab{b}})Liu, Zhou, Xu, and Wang}]{liu2023simplenet}
Liu, Z.; Zhou, Y.; Xu, Y.; and Wang, Z. 2023{\natexlab{b}}.
\newblock Simplenet: A simple network for image anomaly detection and localization.
\newblock In \emph{Proceedings of the IEEE/CVF Conference on Computer Vision and Pattern Recognition}, 20402--20411.

\bibitem[{Mishra et~al.(2021)Mishra, Verk, Fornasier, Piciarelli, and Foresti}]{mishra2021vt}
Mishra, P.; Verk, R.; Fornasier, D.; Piciarelli, C.; and Foresti, G.~L. 2021.
\newblock VT-ADL: A vision transformer network for image anomaly detection and localization.
\newblock In \emph{2021 IEEE 30th International Symposium on Industrial Electronics (ISIE)}, 01--06. IEEE.

\bibitem[{Perlin(1985)}]{perlin}
Perlin, K. 1985.
\newblock An image synthesizer.
\newblock In \emph{Proceedings of the 12th Annual Conference on Computer Graphics and Interactive Techniques}, SIGGRAPH '85, 287–296.
\newblock ISBN 0897911660.

\bibitem[{Roth et~al.(2022)Roth, Pemula, Zepeda, Sch{\"o}lkopf, Brox, and Gehler}]{roth2022towards}
Roth, K.; Pemula, L.; Zepeda, J.; Sch{\"o}lkopf, B.; Brox, T.; and Gehler, P. 2022.
\newblock Towards total recall in industrial anomaly detection.
\newblock In \emph{Proceedings of the IEEE/CVF Conference on Computer Vision and Pattern Recognition}, 14318--14328.

\bibitem[{Rudolph et~al.(2023)Rudolph, Wehrbein, Rosenhahn, and Wandt}]{rudolph2023asymmetric}
Rudolph, M.; Wehrbein, T.; Rosenhahn, B.; and Wandt, B. 2023.
\newblock Asymmetric student-teacher networks for industrial anomaly detection.
\newblock In \emph{Proceedings of the IEEE/CVF winter conference on applications of computer vision}, 2592--2602.

\bibitem[{Salehi et~al.(2021)Salehi, Sadjadi, Baselizadeh, Rohban, and Rabiee}]{salehi2021multiresolution}
Salehi, M.; Sadjadi, N.; Baselizadeh, S.; Rohban, M.~H.; and Rabiee, H.~R. 2021.
\newblock Multiresolution knowledge distillation for anomaly detection.
\newblock In \emph{Proceedings of the IEEE/CVF conference on computer vision and pattern recognition}, 14902--14912.

\bibitem[{Tang et~al.(2020)Tang, Kuo, Lan, Ding, Hsu, and Young}]{tang2020anomaly}
Tang, T.-W.; Kuo, W.-H.; Lan, J.-H.; Ding, C.-F.; Hsu, H.; and Young, H.-T. 2020.
\newblock Anomaly detection neural network with dual auto-encoders GAN and its industrial inspection applications.
\newblock \emph{Sensors}, 20(12): 3336.

\bibitem[{Tien et~al.(2023)Tien, Nguyen, Tran, Huy, Duong, Nguyen, and Truong}]{tien2023revisiting}
Tien, T.~D.; Nguyen, A.~T.; Tran, N.~H.; Huy, T.~D.; Duong, S.; Nguyen, C. D.~T.; and Truong, S.~Q. 2023.
\newblock Revisiting reverse distillation for anomaly detection.
\newblock In \emph{Proceedings of the IEEE/CVF conference on computer vision and pattern recognition}, 24511--24520.

\bibitem[{Wang et~al.(2021)Wang, Han, Ding, and Huang}]{DBLP:conf/bmvc/WangHD021}
Wang, G.; Han, S.; Ding, E.; and Huang, D. 2021.
\newblock Student-Teacher Feature Pyramid Matching for Anomaly Detection.
\newblock In \emph{32nd British Machine Vision Conference 2021, {BMVC} 2021, Online, November 22-25, 2021}, 306. {BMVA} Press.

\bibitem[{Zagoruyko and Komodakis(2016)}]{zagoruyko2016wide}
Zagoruyko, S.; and Komodakis, N. 2016.
\newblock Wide residual networks.
\newblock \emph{arXiv preprint arXiv:1605.07146}.

\bibitem[{Zavrtanik, Kristan, and Sko{\v{c}}aj(2021)}]{zavrtanik2021draem}
Zavrtanik, V.; Kristan, M.; and Sko{\v{c}}aj, D. 2021.
\newblock Draem-a discriminatively trained reconstruction embedding for surface anomaly detection.
\newblock In \emph{Proceedings of the IEEE/CVF International Conference on Computer Vision}, 8330--8339.

\bibitem[{Zhang, Suganuma, and Okatani(2024)}]{zhang2024contextual}
Zhang, J.; Suganuma, M.; and Okatani, T. 2024.
\newblock Contextual affinity distillation for image anomaly detection.
\newblock In \emph{Proceedings of the IEEE/CVF Winter Conference on Applications of Computer Vision}, 149--158.

\bibitem[{Zhang et~al.(2023{\natexlab{a}})Zhang, Li, Li, Dai, Jiang, and Xia}]{zhang2023unsupervised}
Zhang, X.; Li, N.; Li, J.; Dai, T.; Jiang, Y.; and Xia, S.-T. 2023{\natexlab{a}}.
\newblock Unsupervised surface anomaly detection with diffusion probabilistic model.
\newblock In \emph{Proceedings of the IEEE/CVF International Conference on Computer Vision}, 6782--6791.

\bibitem[{Zhang et~al.(2023{\natexlab{b}})Zhang, Li, Li, Huang, Shan, and Chen}]{zhang2023destseg}
Zhang, X.; Li, S.; Li, X.; Huang, P.; Shan, J.; and Chen, T. 2023{\natexlab{b}}.
\newblock Destseg: Segmentation guided denoising student-teacher for anomaly detection.
\newblock In \emph{Proceedings of the IEEE/CVF Conference on Computer Vision and Pattern Recognition}, 3914--3923.

\bibitem[{Zhang, Xu, and Zhou(2024)}]{zhang2024realnet}
Zhang, X.; Xu, M.; and Zhou, X. 2024.
\newblock RealNet: A feature selection network with realistic synthetic anomaly for anomaly detection.
\newblock In \emph{Proceedings of the IEEE/CVF Conference on Computer Vision and Pattern Recognition}, 16699--16708.

\bibitem[{Zhou et~al.(2022)Zhou, He, Liu, Chen, and Chen}]{zhou2022pull}
Zhou, Q.; He, S.; Liu, H.; Chen, T.; and Chen, J. 2022.
\newblock Pull \& push: Leveraging differential knowledge distillation for efficient unsupervised anomaly detection and localization.
\newblock \emph{IEEE Transactions on Circuits and Systems for Video Technology}.

\bibitem[{Zhou et~al.(2024)Zhou, Xu, Song, Shen, and Shen}]{zhou2024msflow}
Zhou, Y.; Xu, X.; Song, J.; Shen, F.; and Shen, H.~T. 2024.
\newblock MSFlow: Multiscale Flow-Based Framework for Unsupervised Anomaly Detection.
\newblock \emph{IEEE Transactions on Neural Networks and Learning Systems}.

\end{thebibliography}


\begin{thebibliography}{13}
\providecommand{\natexlab}[1]{#1}

\bibitem[{Bergmann et~al.(2019)Bergmann, Fauser, Sattlegger, and Steger}]{bergmann2019mvtec}
Bergmann, P.; Fauser, M.; Sattlegger, D.; and Steger, C. 2019.
\newblock MVTec AD--A comprehensive real-world dataset for unsupervised anomaly detection.
\newblock In \emph{Proceedings of the IEEE/CVF conference on computer vision and pattern recognition}, 9592--9600.

\bibitem[{Bergmann et~al.(2020)Bergmann, Fauser, Sattlegger, and Steger}]{bergmann2020uninformed}
Bergmann, P.; Fauser, M.; Sattlegger, D.; and Steger, C. 2020.
\newblock Uninformed students: Student-teacher anomaly detection with discriminative latent embeddings.
\newblock In \emph{Proceedings of the IEEE/CVF conference on computer vision and pattern recognition}, 4183--4192.

\bibitem[{Cimpoi et~al.(2014)Cimpoi, Maji, Kokkinos, Mohamed, and Vedaldi}]{cimpoi2014describing}
Cimpoi, M.; Maji, S.; Kokkinos, I.; Mohamed, S.; and Vedaldi, A. 2014.
\newblock Describing textures in the wild.
\newblock In \emph{Proceedings of the IEEE conference on computer vision and pattern recognition}, 3606--3613.

\bibitem[{Deng and Li(2022)}]{deng2022anomaly}
Deng, H.; and Li, X. 2022.
\newblock Anomaly detection via reverse distillation from one-class embedding.
\newblock In \emph{Proceedings of the IEEE/CVF Conference on Computer Vision and Pattern Recognition}, 9737--9746.

\bibitem[{Gu et~al.(2023)Gu, Liu, Chen, Yi, Zhang, Wang, Wang, Shu, Jiang, and Ma}]{gu2023remembering}
Gu, Z.; Liu, L.; Chen, X.; Yi, R.; Zhang, J.; Wang, Y.; Wang, C.; Shu, A.; Jiang, G.; and Ma, L. 2023.
\newblock Remembering Normality: Memory-guided Knowledge Distillation for Unsupervised Anomaly Detection.
\newblock In \emph{Proceedings of the IEEE/CVF International Conference on Computer Vision}, 16401--16409.

\bibitem[{Jezek et~al.(2021)Jezek, Jonak, Burget, Dvorak, and Skotak}]{jezek2021deep}
Jezek, S.; Jonak, M.; Burget, R.; Dvorak, P.; and Skotak, M. 2021.
\newblock Deep learning-based defect detection of metal parts: evaluating current methods in complex conditions.
\newblock In \emph{2021 13th International congress on ultra modern telecommunications and control systems and workshops (ICUMT)}, 66--71. IEEE.

\bibitem[{Mishra et~al.(2021)Mishra, Verk, Fornasier, Piciarelli, and Foresti}]{mishra2021vt}
Mishra, P.; Verk, R.; Fornasier, D.; Piciarelli, C.; and Foresti, G.~L. 2021.
\newblock VT-ADL: A vision transformer network for image anomaly detection and localization.
\newblock In \emph{2021 IEEE 30th International Symposium on Industrial Electronics (ISIE)}, 01--06. IEEE.

\bibitem[{Perlin(1985)}]{perlin}
Perlin, K. 1985.
\newblock An image synthesizer.
\newblock In \emph{Proceedings of the 12th Annual Conference on Computer Graphics and Interactive Techniques}, SIGGRAPH '85, 287–296.
\newblock ISBN 0897911660.

\bibitem[{Tien et~al.(2023)Tien, Nguyen, Tran, Huy, Duong, Nguyen, and Truong}]{tien2023revisiting}
Tien, T.~D.; Nguyen, A.~T.; Tran, N.~H.; Huy, T.~D.; Duong, S.; Nguyen, C. D.~T.; and Truong, S.~Q. 2023.
\newblock Revisiting reverse distillation for anomaly detection.
\newblock In \emph{Proceedings of the IEEE/CVF conference on computer vision and pattern recognition}, 24511--24520.

\bibitem[{Zagoruyko and Komodakis(2016)}]{zagoruyko2016wide}
Zagoruyko, S.; and Komodakis, N. 2016.
\newblock Wide residual networks.
\newblock \emph{arXiv preprint arXiv:1605.07146}.

\bibitem[{Zavrtanik, Kristan, and Sko{\v{c}}aj(2021)}]{zavrtanik2021draem}
Zavrtanik, V.; Kristan, M.; and Sko{\v{c}}aj, D. 2021.
\newblock Draem-a discriminatively trained reconstruction embedding for surface anomaly detection.
\newblock In \emph{Proceedings of the IEEE/CVF International Conference on Computer Vision}, 8330--8339.

\bibitem[{Zhang et~al.(2023)Zhang, Li, Li, Huang, Shan, and Chen}]{zhang2023destseg}
Zhang, X.; Li, S.; Li, X.; Huang, P.; Shan, J.; and Chen, T. 2023.
\newblock Destseg: Segmentation guided denoising student-teacher for anomaly detection.
\newblock In \emph{Proceedings of the IEEE/CVF Conference on Computer Vision and Pattern Recognition}, 3914--3923.

\bibitem[{Zou et~al.(2022)Zou, Jeong, Pemula, Zhang, and Dabeer}]{zou2022spot}
Zou, Y.; Jeong, J.; Pemula, L.; Zhang, D.; and Dabeer, O. 2022.
\newblock Spot-the-difference self-supervised pre-training for anomaly detection and segmentation.
\newblock In \emph{European Conference on Computer Vision}, 392--408. Springer.

\end{thebibliography}

\clearpage
\twocolumn[
    \begin{@twocolumnfalse}
        \LARGE{\centering \textbf{Supplementary material:} \\[0.5em]
Unlocking the Potential of Reverse Distillation for Anomaly Detection\\[1em]}
    \end{@twocolumnfalse}
]

\appendix

\section{Details of Guided Information Injection} \label{Sec3-3}

To selectively introduce the normal features from the teacher encoder into the student decoder and aid in detailed reconstruction, we propose Guided Information Injection (GII). In this section, the specific details of GII are provided, as shown in Algorithm~\ref{alg:algorithm}. $\odot$ represents the element-wise multiplication operation. The resulting $F_{S_{GII}}^{i+1}$ will replace the original $F_S^{i+1}$ in the RD architecture and be fed into $S^i$ to reconstruct the feature $F_S^{i}$.

\begin{algorithm}[h]
\caption{Guided Information Injection defore $S^i$}
\label{alg:algorithm}
\textbf{Input}: teacher features $F_T^{i}$ and $F_T^{i+1}$, student feature $F_S^{i+1}$\\
\textbf{Output}: output feature $F_{S_{GII}}^{i+1}$
\begin{algorithmic}[1] 
\STATE ${F_T^i}^{'} = \mathrm{Conv_{1 \times 1}}(\mathrm{DownSample}(F_T^{i}))$\\
\STATE $F_{T_{fuse}}^{i+1} = \mathrm{Conv_{3 \times 3}}({F_T^i}^{'} + F_T^{i+1})$
\STATE $\mathrm{Sim}^{i+1} = \frac{F_T^{i+1}·F_S^{i+1}}{\Vert F_T^{i+1}\Vert\Vert F_S^{i+1}\Vert}$ \\
\STATE $F_{S_{SA}}^{i+1} = \mathrm{Conv_{3 \times 3}}(F_{T_{fuse}}^{i+1} \odot \mathrm{Sim}  + F_S^{i+1} \odot (1-\mathrm{Sim}))$
\STATE $F_{S_{GII}}^{i+1} = \mathrm{Conv_{3 \times 3}}(\{F_{S_{SA}}^{i+1}, F_{S}^{i+1}\})$
\end{algorithmic}
\end{algorithm}

\section{Details of Anomaly Synthesis}

\begin{figure*}
\centering
\includegraphics[width=\linewidth]{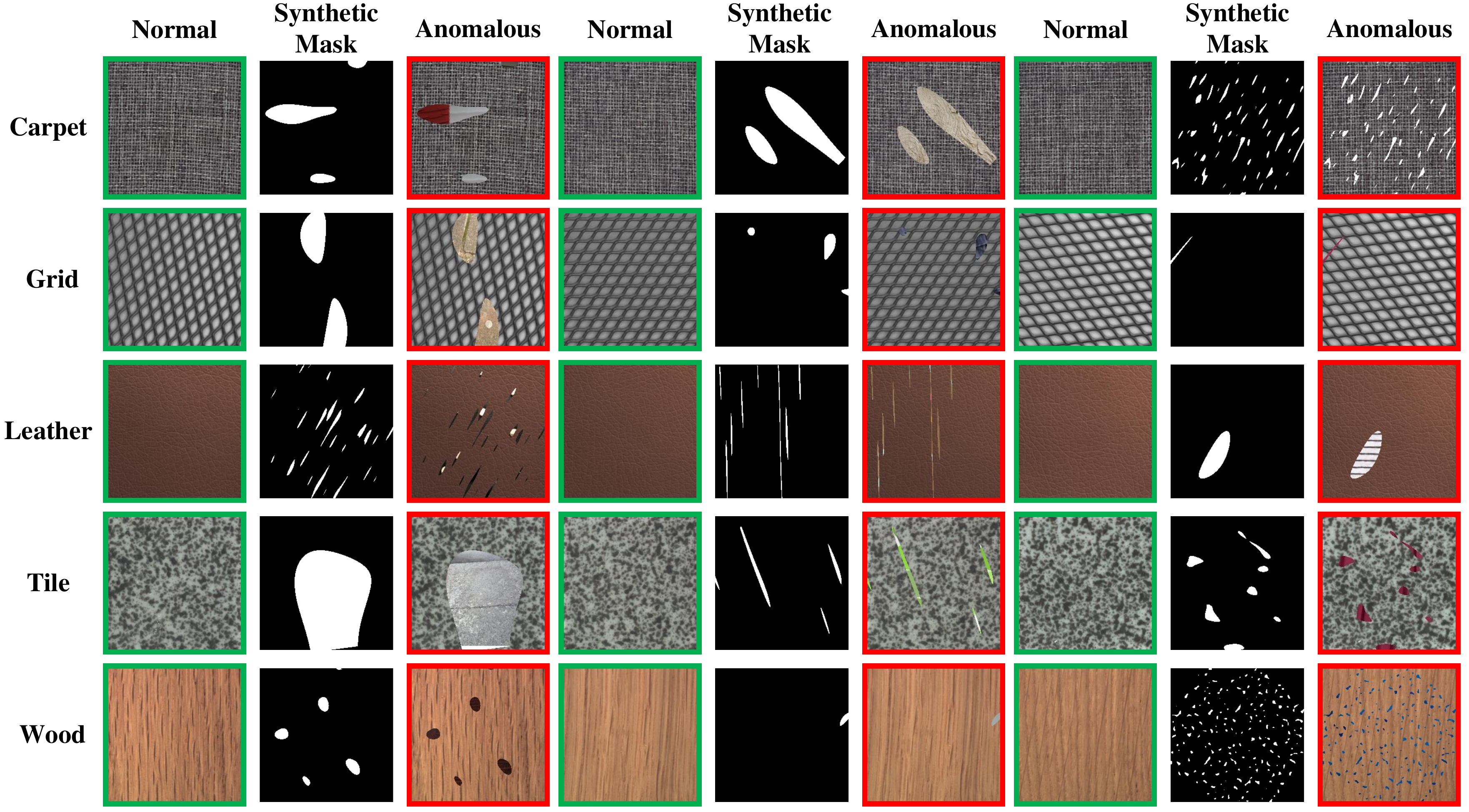} 
\caption{Anomaly synthesis of texture images in MVTec AD.
}
\label{fig_t}
\end{figure*}

\begin{figure*}
\centering
\includegraphics[width=\linewidth]{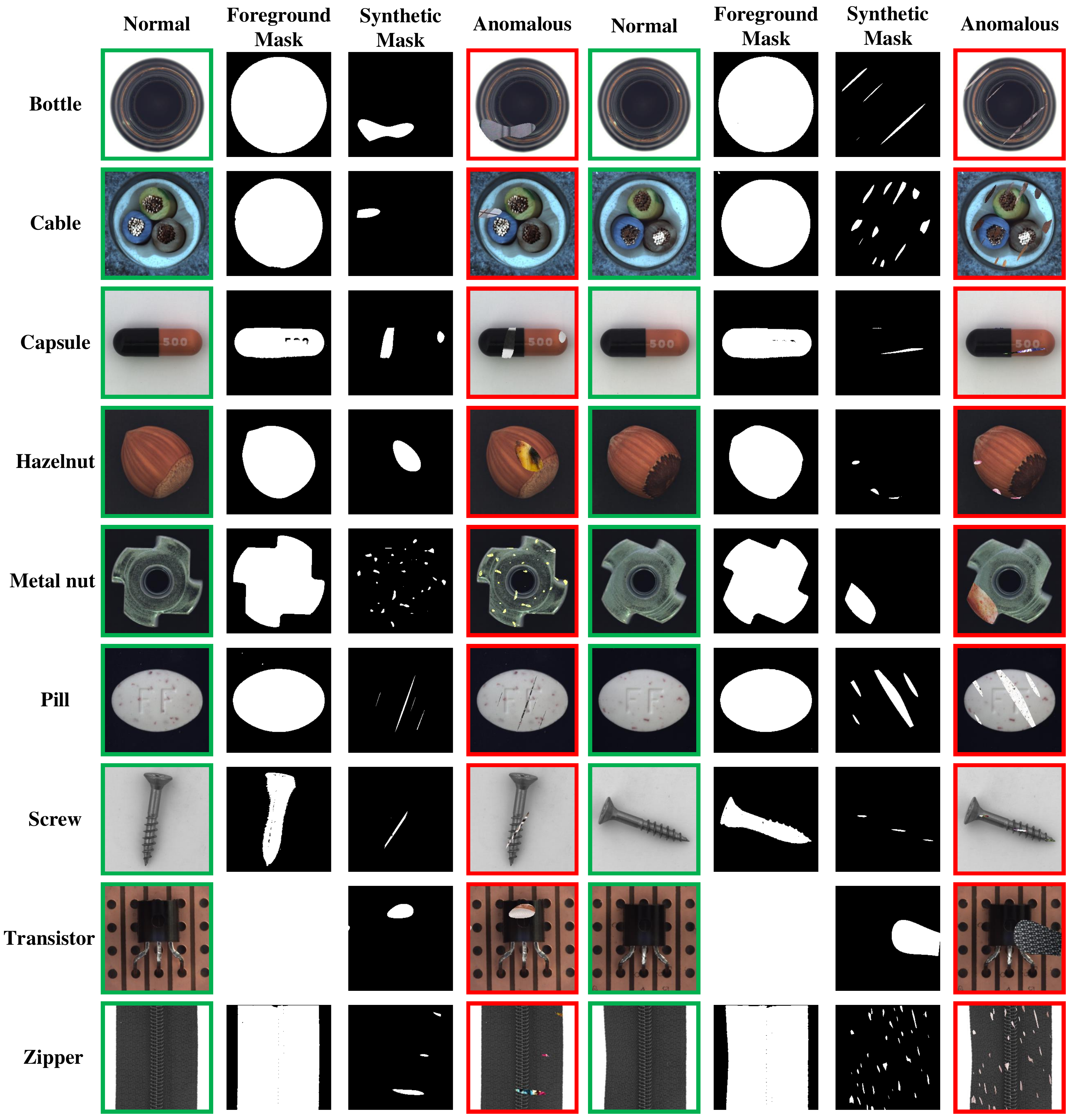} 
\caption{Anomaly synthesis of object images in MVTec AD.
}
\label{fig_o}
\end{figure*}

Our method relies on synthetic anomalies for training the expert network to guide both the teacher and the student. The anomaly synthesis approach used in our method is based on DRÆM \citesupp{zavrtanik2021draem}. First, generating the synthetic masks: the random two-dimensional Perlin noise \citesupp{perlin} is binarized to obtain the synthetic anomaly masks. Then, using $M$, the images from external texture dataset DTD \citesupp{cimpoi2014describing} are overlaid onto normal images to create synthetic anomaly images. 
The corresponding synthetic anomalous image $I_a$ for a given normal image $I_a$ is expressed as
\begin{equation}
\fontsize{9}{10}\selectfont
 I_a = (1-M_{syn}) \odot I_n + (1-\beta)(M_{syn} \odot I_n) + \beta(M_{syn} \odot I_t)
\end{equation}
where $M_{syn}$ is the corresponding synthetic mask, $\beta$ is the opacity parameter chosen between $[0.15, 1]$ following DeSTSeg \citesupp{zhang2023destseg}, and $I_t$ is an image randomly obtained from the external dataset.

For object categories, we make a slight modification to previous approach to let synthetic anomalies be more similar to actual ones. Specifically, we restrict the anomaly regions to the object's foreground, which is achieved by using foreground masks to exclude background regions when generating the synthetic anomaly masks. The foreground masks are simply obtained by binarizing the images and setting a threshold.

We provide some examples of synthetic anomalies, with texture categories shown in Figure \ref{fig_t} and object categories in Figure \ref{fig_o}.

\section{Experimental Setup}
\subsection{Details of Datasets}
\subsubsection{MVTec AD}

MVTec AD \citesupp{bergmann2019mvtec} is a commonly used benchmark for unsupervised AD. It consists of 15 categories of industrial images, including 5 texture categories and 10 object categories. The training set contains 3629 normal images, while the test set includes both normal and anomalous images, with a total of 1725 images.

\subsubsection{MPDD}

MPDD \citesupp{jezek2021deep} is developed for the defect detection of metal parts, reflecting anomalies that occur on human-operated production lines. It includes a total of six categories of metal part images. The training set contains 888 normal images, while the test set includes 176 normal images and 282 defect images.

\subsubsection{BTAD}

BTAD \citesupp{mishra2021vt} dataset comprises 3 categories of industrial product images, with a total of 1,799 training images and 736 test images. Some misclassified images are present in the original BTAD dataset, which have been removed before our experiments.

\subsubsection{VisA}
VisA \citesupp{zou2022spot} has 12 categories of images, which are grouped into three main types: Complex structure, Multiple instances, and Single instance. A total of 10,821 images are included in this dataset, with 9,621 normal images and 1,200 anomalous images. For unsupervised AD task, we first divide VisA according to unsupervised standards, where the training set contains only normal images, and the test set includes both normal and anomalous images.

\subsection{Additional Implementation Details}

Consistent with RD, we used the Adam optimizer with $\beta=(0.5, 0.999)$. The backbone network we employ is WideResNet50 \citesupp{zagoruyko2016wide}, and the output anomaly maps are computed using the first three stages of the teacher and student networks. During inference, the ground truth anomaly masks are also resized to $256 \times 256$. The code is implemented in PyTorch 2.0, and all experiments are conducted on a single NVIDIA GeForce RTX 3090 GPU.

\subsection{Evaluation Metrics}

We introduce AUC and AP metrics for anomaly detection and localization. AUC measures the model's ability to distinguish between normal and anomalous samples across all thresholds, while AP emphasizes the balance between precision and recall, making it more suitable for imbalanced categories. For anomaly localization, considering that the size of anomalies may affect AUC, we also introduce PRO \citesupp{bergmann2020uninformed}. PRO assesses the accuracy of anomaly localization by measuring the overlap between predicted and ground truth anomalous regions, treating anomalies of all sizes equally.

Throughout the experimental results, we use \textbf{I-AUC} and \textbf{I-AP} to refer to image-level AUC and AP metrics, and \textbf{P-AUC}, \textbf{P-AP}, and \textbf{P-PRO} to refer to pixel-level AUC, AP, and PRO metrics.

\section{Complete Anomaly Detection Results}

This section presents the complete anomaly detection results for each category. Tables \ref{tab:mvtec_detection}, \ref{tab:mpdd}, and \ref{tab:btad} respectively show the detection results on MVTec AD, MPDD, and BTAD datasets. Overall, our method achieves state-of-the-art results on AP compared with other advanced RD-based methods on MVTec AD and BTAD datasets. Additionally, our method achieves the best detection performance in terms of AUC and AP across many categories. Although the anomaly detection performance on MPDD fall short of MemKD \citesupp{gu2023remembering}, they still surpass the baseline method RD \citesupp{deng2022anomaly}. Furthermore, in some categories such as Connector, Metal Plates, and Tubes, our method either exceeds or matches the performance of other RD-based competitors.

\begin{table*}
\fontsize{9}{10}\selectfont
\centering
\begin{tabular}{cc|ccc|ccccc}
    \toprule
    \multicolumn{2}{c|}{\multirow{2}{*}{Category}} 
    & \multicolumn{3}{c|}{Forward Distillation} &  \multicolumn{5}{c}{Reverse Distillation} \\
 & &STPM & DeSTSeg &  HypAD&  RD & RD++ & THFR & MemKD&	Ours
\\
    \midrule
   \multirow{6}{*}{\rotatebox[origin=c]{90}{Textures}}  & Carpet  & -& 98.9/- &-& \textbf{100}/\textbf{100}& \textbf{100}/\textbf{100} & 99.8/-  & 99.6/- &   99.5/99.9               \\
              & Grid    & - & 99.7/- &  - & \textbf{100}/\textbf{100} & \textbf{100}/\textbf{100} & \textbf{100}/- & \textbf{100}/-  &  \textbf{100}/\textbf{100}                 \\
              & Leather & - &\textbf{100}/- & - & \textbf{100}/\textbf{100}& \textbf{100}/\textbf{100} &  \textbf{100}/- & \textbf{100}/- & \textbf{100}/\textbf{100}   \\
              & Tile    & - & \textbf{100}/- & - & 99.4/99.8& 99.8/99.9 & 99.3/- & \textbf{100}/- & \textbf{100}/\textbf{100}                \\
              & Wood    & - & 97.1/- & - & 99.2/99.8& 99.3/99.8 & 99.2/- & 99.5/- & \textbf{99.6}/\textbf{99.9}            \\
              \cmidrule(r){2-10} 
              & Average  & - & 99.1/- & - & 99.7/99.9& 99.8/99.9 & 99.7/- & \textbf{99.8}/- & \textbf{99.8}/\textbf{100}\\
              \midrule
 \multirow{11}{*}{\rotatebox[origin=c]{90}{Objects}}  & Bottle    & - &  \textbf{100}/- &-& \textbf{100}/\textbf{100}& \textbf{100}/\textbf{100} & \textbf{100}/- & \textbf{100}/- &  \textbf{100}/\textbf{100}            \\
              & Cable     & - & 97.8/-  & - & 97.1/98.5& 97.8/98.9 & \textbf{99.2}/- & \textbf{99.2}/- & 98.6/\textbf{99.2}                \\
              & Capsule & - & 97.0/- & - & 98.0/99.6& 97.5/99.4 & 97.5/- & \textbf{98.8}/-  &  98.5/\textbf{99.7}          \\
              & Hazelnut & - & 99.9/-  & - & \textbf{100}/\textbf{100}& \textbf{100}/\textbf{100} & \textbf{100}/- & \textbf{100}/- &  \textbf{100}/\textbf{100}           \\
              & Metal\_nut & - & 99.5/- & - & 98.6/99.7& \textbf{100}/\textbf{100} & \textbf{100}/- & \textbf{100}/- & 99.7/99.9               \\
              & Pill      & - & 97.2/- & - & 96.5/99.4& 97.3/99.5 & 97.8/- & 98.3/- & \textbf{98.5}/\textbf{99.7}             \\
              & Screw   & - & 93.6/- & - & 98.7/\textbf{99.6}& 98.5/99.5 & 97.1/- & \textbf{99.1}/- &  95.8/98.5     \\
              & Toothbrush & - & 99.9/- & - & 98.9/99.6& 98.9/99.6 & \textbf{100}/- & \textbf{100}/- & \textbf{100}/\textbf{100}         \\
              & Transistor & - & 98.5/- & - & 97.1/97.6& 98.3/98.1 &  99.7/- & \textbf{100}/- &  99.9/\textbf{99.9}        \\
              & Zipper    & - & \textbf{100}/- & - & 98.1/99.4& 97.4/99.2 & 97.7/- & 99.3/- &  98.3/\textbf{99.5}           \\
              \cmidrule(r){2-10} 
              & Average   & - & 98.3/- & - & 98.3/99.3& 98.6/99.4 & 98.9/- & \textbf{99.5}/- & 98.9/\textbf{99.6} \\
              \midrule
 \multicolumn{2}{c|}{Total Average} & 95.5/- & 98.6/- & 99.2/99.5& 98.8/99.5& 99.0/99.6 & 99.2/- & \textbf{99.6}/- &  99.2/\textbf{99.7}      \\
  \bottomrule
\end{tabular}
\caption{Image-level anomaly detection results I-AUC/I-AP (\%) on MVTec AD with the best in bold.}
\label{tab:mvtec_detection}
\end{table*}

\begin{table}
\fontsize{9}{10}\selectfont
\centering
  \begin{tabular}{c|cccc}
    \toprule
   Category &  RD & RD++ & MemKD &	Ours \\
\midrule
Bracket Black    & 90.3/93.7  & 89.5/92.8  & \textbf{91.2}/\textbf{94.4} & 89.9/94.0 \\
Bracket Brown   & 92.5/95.1  & 92.4/95.6  & \textbf{95.2}/\textbf{97.3} & 91.9/95.2 \\
Bracket White  & 89.2/91.7  &  90.3/91.9  & \textbf{92.7}/\textbf{93.8} &  88.6/92.4  \\
Connector  &  99.8/99.5 &  99.5/99.0  &  \textbf{100}/\textbf{100} & \textbf{100}/\textbf{100}  \\
Metal Plate  & \textbf{100}/\textbf{100}  &  \textbf{100}/\textbf{100}  & \textbf{100}/\textbf{100} & \textbf{100}/\textbf{100}   \\
Tubes  &  96.7/98.7 &  92.6/97.1  & 96.2/98.5 & \textbf{97.6}/\textbf{99.0}   \\
Average  & 94.8/96.5 &  94.1/96.1 &  \textbf{95.4}/\textbf{97.3} & 94.7/96.8  \\
  \bottomrule
\end{tabular}
\caption{Image-level anomaly detection results I-AUC/I-AP (\%) on MPDD with the best in bold.}
\label{tab:mpdd}
\end{table}

\begin{table}
\fontsize{9}{10}\selectfont
\centering
  \begin{tabular}{c|ccc}
    \toprule
   Category &  RD & RD++ &	Ours \\
\midrule
Class 01    & \textbf{97.9}/\textbf{99.3}  &  97.3/98.9 & 96.0/98.2   \\
Class 02   & 86.0/97.7  &  \textbf{87.3}/\textbf{97.9} &  85.8/97.7  \\
Class 03  &  99.7/94.2 &  99.7/95.9  &   \textbf{99.8}/\textbf{97.0}   \\
Average  &  94.5/97.1 &  \textbf{94.8}/\textbf{97.6} &   93.9/\textbf{97.6}   \\
  \bottomrule
\end{tabular}
\caption{Image-level anomaly detection results I-AUC/I-AP (\%) on BTAD with the best in bold.}
\label{tab:btad}
\end{table}

\section{Experimental Results on VisA}

This section shows the anomaly detection and localization results of our method on the VisA dataset, as shown in Table \ref{tab:visa}. The methods we compare with are RD-based, including RD \citesupp{deng2022anomaly} and MemKD \citesupp{gu2023remembering}. As is shown in the table, our proposed method also outperforms other RD-based SOTA methods on VisA in P-AUC and P-AP, achieving 99.1\% and 47.7\%.

To clarify, we first downsample the ground truth masks to $256 \times 256$ and then test the model using the downsampled anomaly masks. In the VisA dataset, some anomalies are too small to be distinguishable after downsampling, which may result in deviations. To ensure a fairer comparison, we re-implement RD on VisA using the same testing approach with our method.

\begin{table*}
\fontsize{9}{10}\selectfont
\setlength{\tabcolsep}{0.8mm}
\centering
\begin{tabular}{cc|ccccc|ccccc|ccccc}
    \toprule
    \multicolumn{2}{c|}{\multirow{2}{*}{Category}} 
    & \multicolumn{5}{c|}{MemKD} &  \multicolumn{5}{c|}{RD}  & \multicolumn{5}{c}{Ours}\\
 & &I-AUC & I-AP & P-AUC& P-AP & P-PRO &I-AUC & I-AP & P-AUC& P-AP & P-PRO &I-AUC & I-AP & P-AUC& P-AP & P-PRO
\\
    \midrule
   \multirow{4}{*}{\rotatebox[origin=c]{90}{\makecell[c]{Complex \\structure}}}  & PCB1  &96.9 & \textbf{99.7}& \textbf{99.8}& 82.1& \textbf{96.9}& \textbf{97.8}& 97.7& \textbf{99.8}& 82.3& \textbf{96.9}& 97.4& 97.3& \textbf{99.8}& \textbf{84.5}& 96.2 \\
              & PCB2   & \textbf{98.0}& 94.8& 96.0& 25.2& \textbf{94.9}& 97.8& \textbf{97.6}& 98.9& \textbf{26.3}& 93.3& 96.3& 96.5& \textbf{99.0}& 23.1& 92.9 \\
              & PCB3   & \textbf{97.8}& \textbf{99.1}& 99.3& 35.6& \textbf{96.6}& 96.9& 97.1& \textbf{99.4}& 35.3& 95.2& 97.5& 97.8& \textbf{99.4}& \textbf{44.1}& 95.6 \\
              & PCB4   & 99.8& 98.6& 98.6& 44.3& \textbf{99.9}& \textbf{99.9}& \textbf{99.9}& 98.7& 44.3& 91.0& \textbf{99.9}& \textbf{99.9}& \textbf{99.1}& \textbf{47.7}& 92.1 \\
              \midrule
    \multirow{4}{*}{\rotatebox[origin=c]{90}{\makecell[c]{Multiple \\instances}}}  & Capsules  & \textbf{94.7}& \textbf{99.0}& 99.2& 58.2& 88.2& 92.1& 95.2& \textbf{99.4}& \textbf{60.8}& \textbf{96.7}& 91.4& 95.2& \textbf{99.4}& 57.8& 95.3 \\
              & Candle   & 95.9& \textbf{99.1}& 99.0& 23.1& 93.8& 95.0& 95.4& 99.1& 23.2& \textbf{95.6}& \textbf{97.7}& 98.0& \textbf{99.3}& \textbf{32.9}& 95.0 \\
              & Macaroni1   & \textbf{98.0}& \textbf{99.6}& 99.6& 23.2& 92.7& 95.9& 94.4& \textbf{99.8}& \textbf{21.9}& 98.3& 95.7& 94.1& \textbf{99.8}& 21.7& \textbf{98.7} \\
              & Macaroni2   & \textbf{92.0}& \textbf{99.2}& 99.5& 13.0& 84.8& 89.2& 86.5& \textbf{99.8}& 12.8& 98.9& 90.1& 87.2& 99.7& \textbf{13.1}& \textbf{99.0} \\
              \midrule          
    \multirow{4}{*}{\rotatebox[origin=c]{90}{\makecell[c]{Single \\instance}}}  & Cashew  & \textbf{99.4}& \textbf{98.7}& 96.6& 58.2& \textbf{97.5}& 96.8& 98.6& 96.2& 54.5& 94.5& 96.6& 98.4& \textbf{98.2}& \textbf{63.5}& 95.4 \\
              & Chewing gum   & \textbf{99.8}& 99.1& 98.6& 60.3& \textbf{98.8}& 97.1& 98.6& \textbf{99.3}& 66.7& 92.9& \textbf{99.5}& 99.7& 99.0& \textbf{76.3}& 89.4 \\
              & Fryum   & \textbf{98.8}& 97.0& 96.9& \textbf{49.3}& \textbf{96.6}& 96.7& 98.5& 96.9& 48.4& 93.9& 96.7& \textbf{98.6}& \textbf{97.0}& 48.3& 94.7 \\
              & Pipe fryum   & \textbf{100}& 99.2& 99.2& 56.2& \textbf{99.0}& 99.6& \textbf{99.8}& 99.1& 54.2& 97.2& 99.6& \textbf{99.8}& \textbf{99.3}& \textbf{58.9}& 97.1 \\
              \midrule   
 \multicolumn{2}{c|}{Total Average} &\textbf{97.6} & \textbf{98.6}& 98.4& 44.1& 94.9& 96.2& 96.6& 98.9& 44.2& \textbf{95.4}& 96.5& 96.9& \textbf{99.1}& \textbf{47.7}& 95.1 \\
  \bottomrule
\end{tabular}
\caption{Anomaly detection and localization results (\%) on VisA with the best in bold.}
\label{tab:visa}
\end{table*}

\section{More Ablation Results}

\subsubsection{Ablation Study on Network Composition}

\begin{figure*}
\centering
\includegraphics[width=\linewidth]{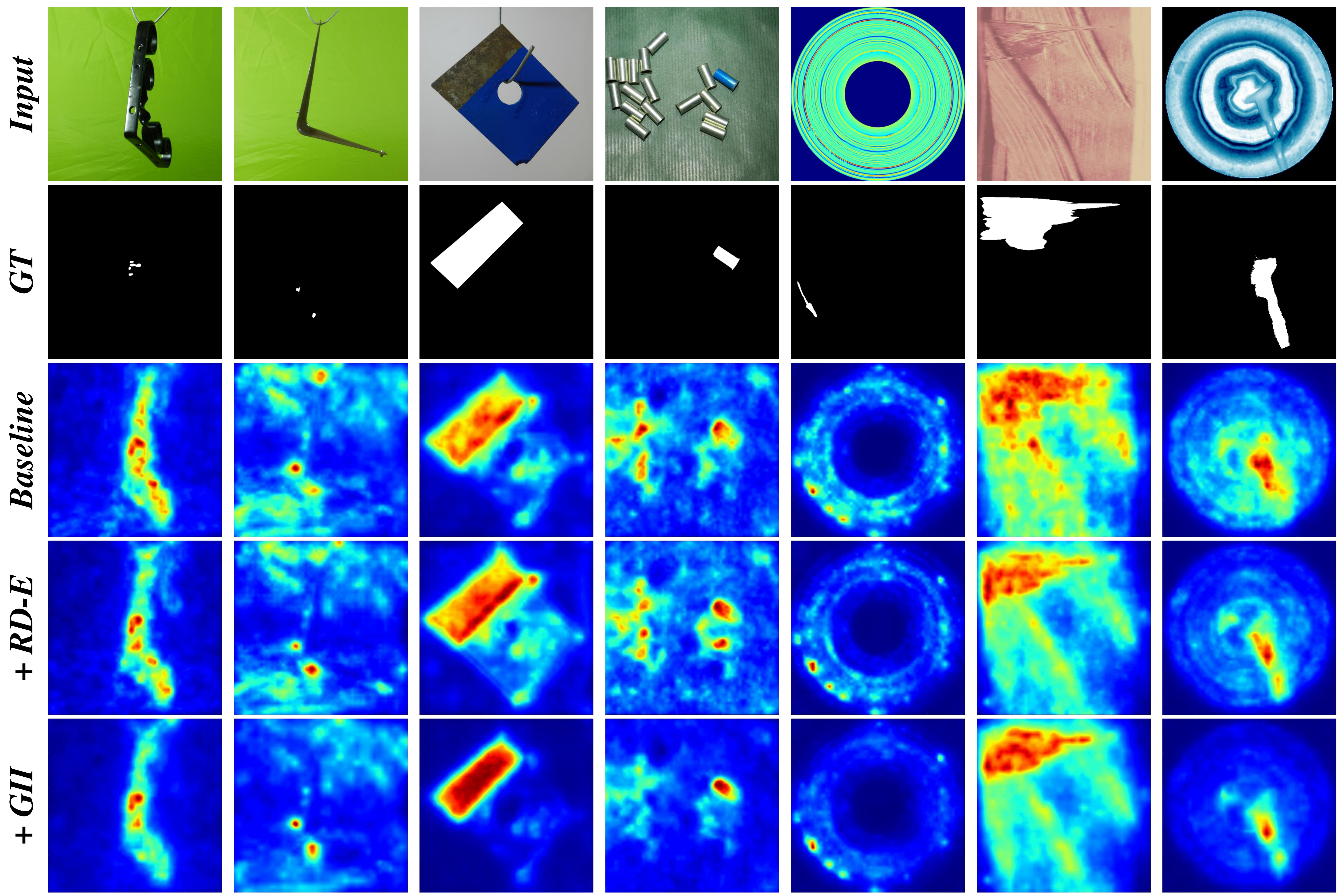} 
\caption{Visualization of ablation study on network composition on MPDD and BTAD. 
From top to bottom: the input image, the ground truth masks, the output anomaly maps of Baseline (RD), Baseline+RD-E, and Baseline+RD-E+GII (Ours).
}
\label{fig_vis_mpddbtad}
\end{figure*}

This section provides the qualitative ablation studies on the network composition conducted on MPDD and BTAD datasets, as in Figure \ref{fig_vis_mpddbtad}.

\subsubsection{Ablation Study on Guided Information Injection}

We extend the ablation study on Guided Information Injection to MPDD and BTAD datasets. Table \ref{tab:ablation3} shows the anomaly localization results. The addition of skip connection leads to improvements across all three localization metrics on both datasets. Additionally, after applying the proposed similarity attention, both AUC and AP metrics are further improved while PRO remains stable.

\begin{table}
\fontsize{9}{10}\selectfont
\setlength{\tabcolsep}{1.0mm}
\centering
  \begin{tabular}{cc|ccc|ccc}
    \toprule
     &  & \multicolumn{3}{c|}{MPDD} & \multicolumn{3}{c}{BTAD}\\ 
  &   &  P-AUC & P-AP & P-PRO &  P-AUC & P-AP & P-PRO\\
    \midrule
\multicolumn{2}{c|}{w/o GII} & 98.76& 45.75& 96.06 & 97.97 & 63.63& 77.78 \\
 \midrule
\multirow{2}{*}{w/ GII}& + SC &99.08 & 51.45& 97.39 & 98.08 &64.77 & 78.49 \\
&+ SA & 99.14 & 52.46 & 97.25& 98.11 & 65.24 & 78.48 \\
  \bottomrule
\end{tabular}
\caption{Ablation study results (\%) of 
GII
on MPDD and BTAD. (SC: Naive skip connection. SA: Similarity attention.)}
\label{tab:ablation3}
\end{table}

\section{More Visualizations}
\subsection{Visual Analysis of Detection Errors in RD} 

\begin{figure*}
\centering
\includegraphics[width=\linewidth]{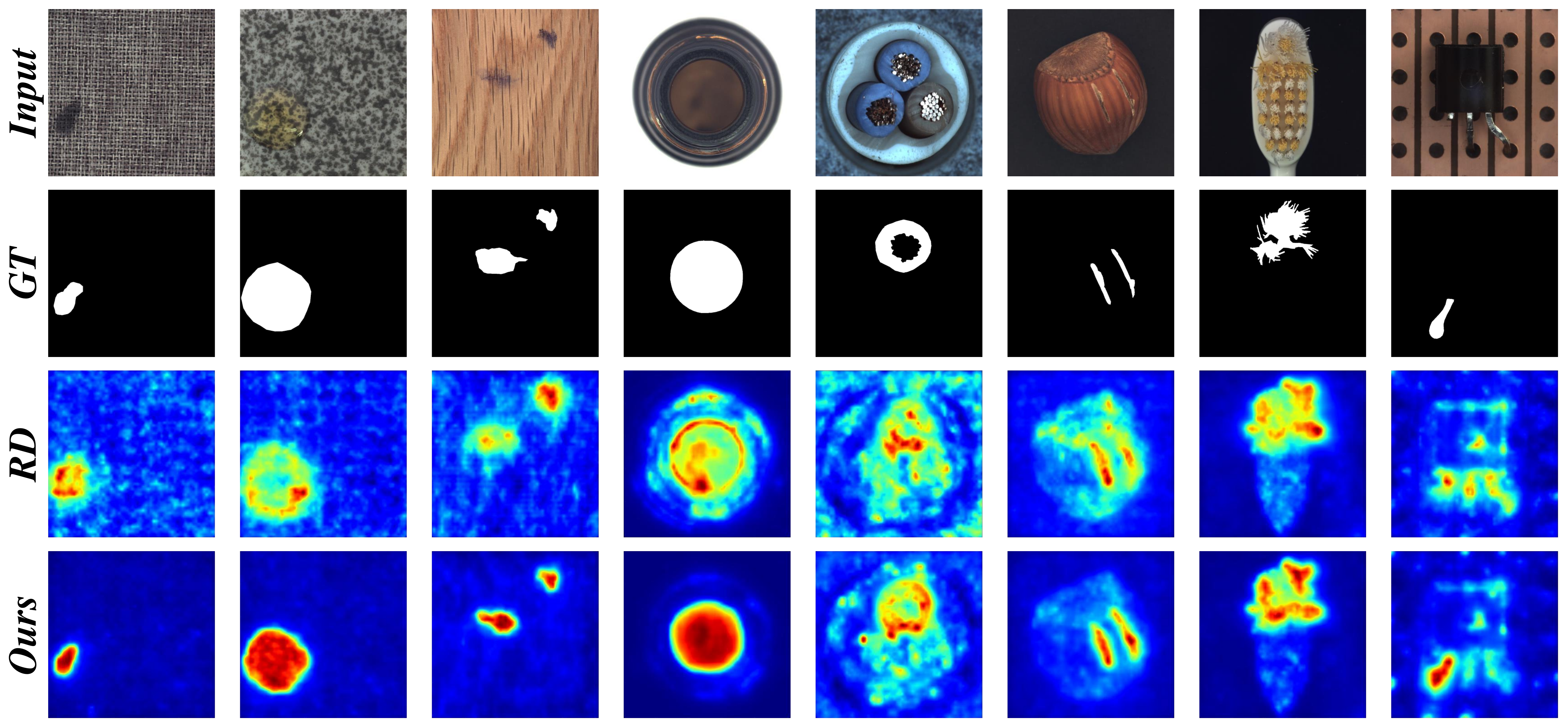} 
\caption{Missed detections of RD.
}
\label{fig_miss}
\end{figure*}

\begin{figure*}
\centering
\includegraphics[width=\linewidth]{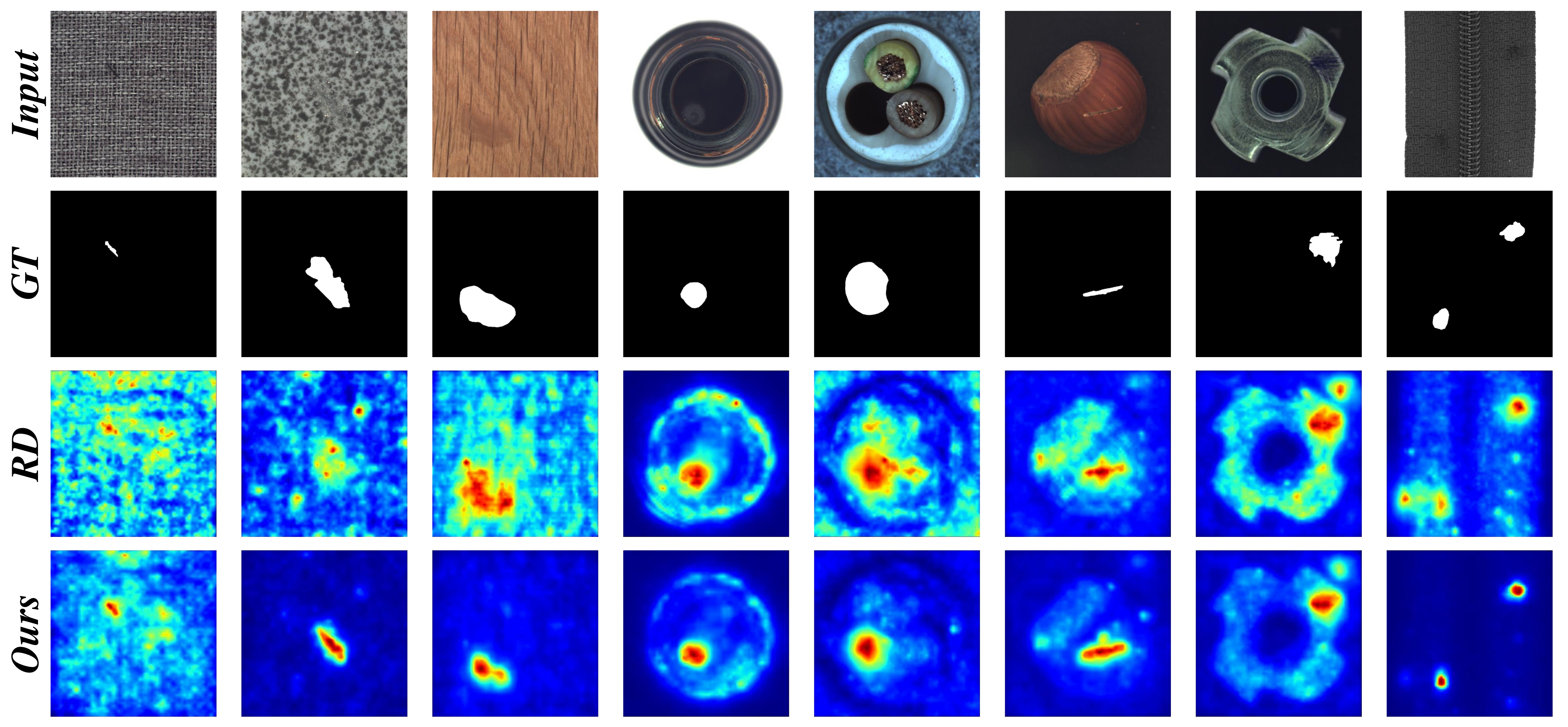} 
\caption{False positives of RD.
}
\label{fig_false}
\end{figure*}

Figures \ref{fig_miss} and \ref{fig_false} respectively illustrate examples of missed detections and false positives by RD on MVTec AD. As shown in Figure \ref{fig_miss}, missed detections often occur in the central region of large-scale anomalies. Our method effectively addresses this issue, yielding more complete anomaly regions. In Figure \ref{fig_false}, RD tends to produce false positives in region like textures and backgrounds, mistakenly identifying normal regions as anomalies. Our proposed method significantly reduces noise in the anomaly maps, mitigating these false positives.

\subsection{Qualitative comparisons on more datasets}

\begin{figure*}
\centering
\includegraphics[width=\linewidth]{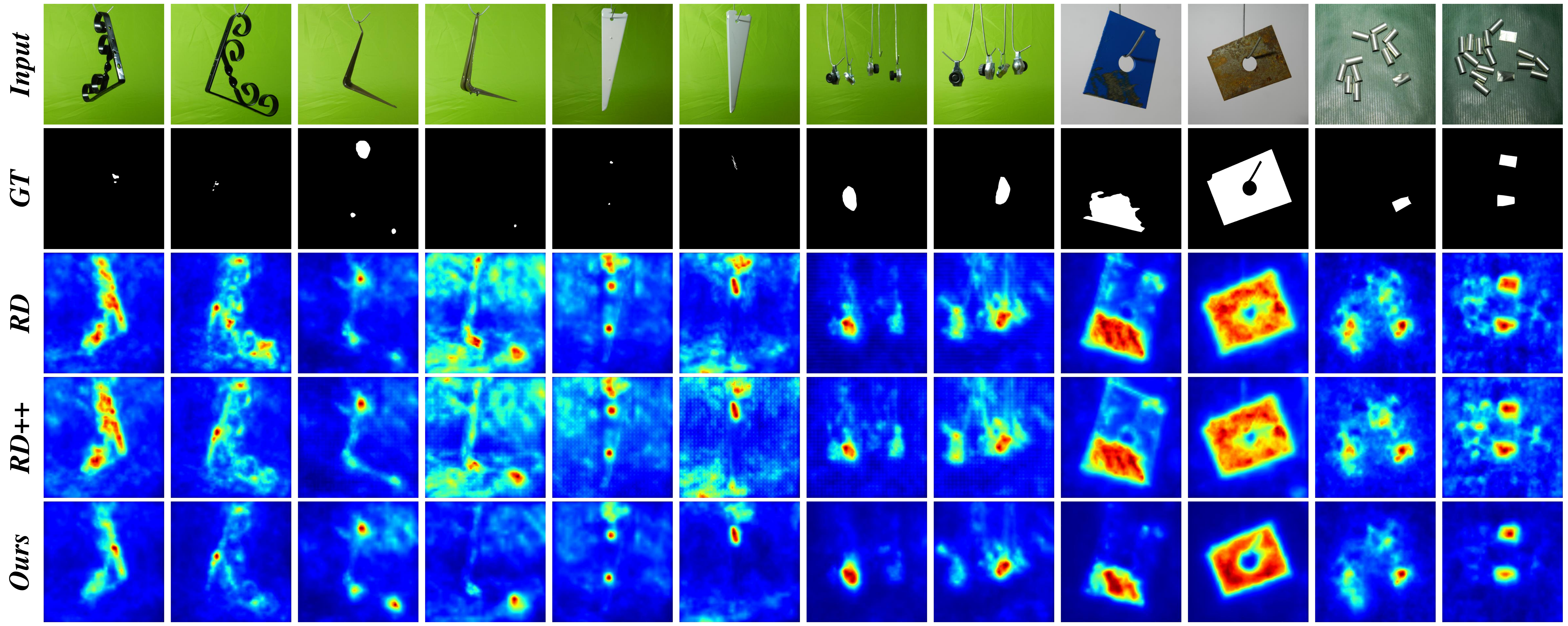} 
\caption{Visualization of anomaly localization on MPDD.
}
\label{fig_mpdd}
\end{figure*}

\begin{figure*}
\centering
\includegraphics[width=\linewidth]{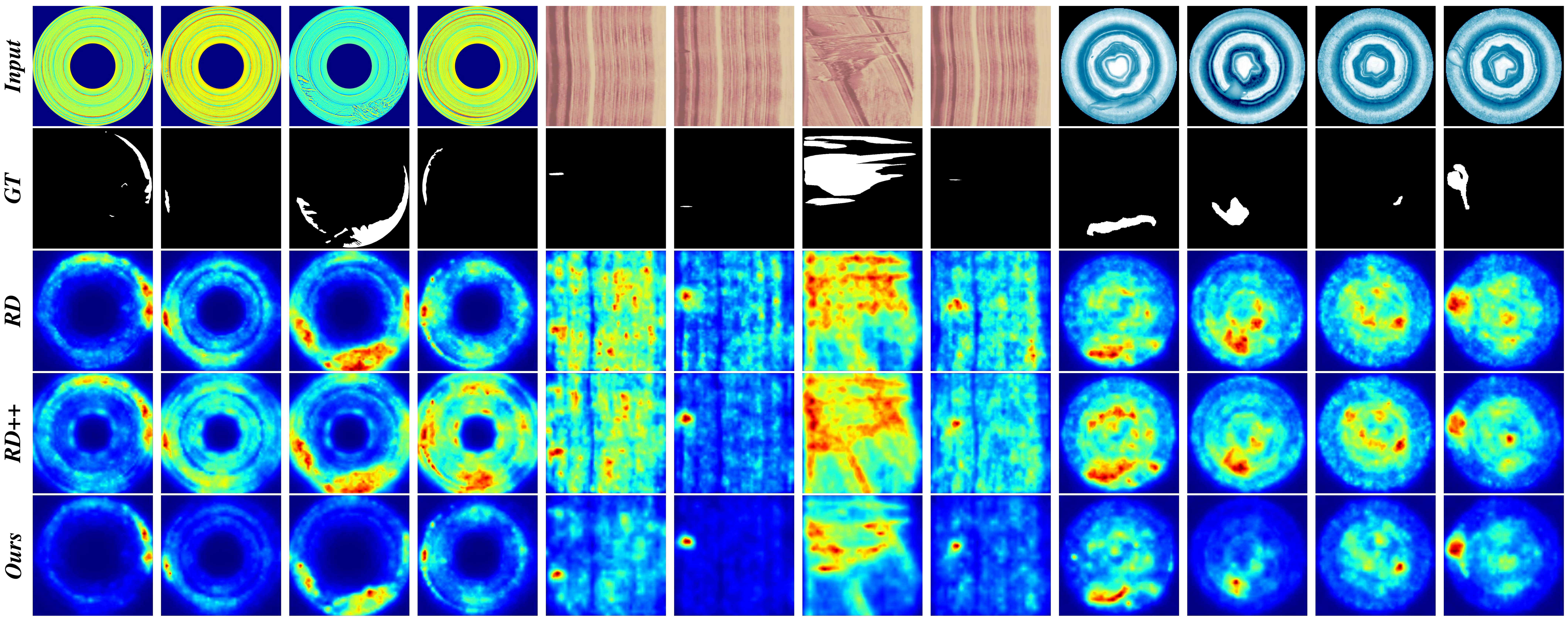} 
\caption{Visualization of anomaly localization on BTAD.
}
\label{fig_btad}
\end{figure*}

\begin{figure*}
\centering
\includegraphics[width=\linewidth]{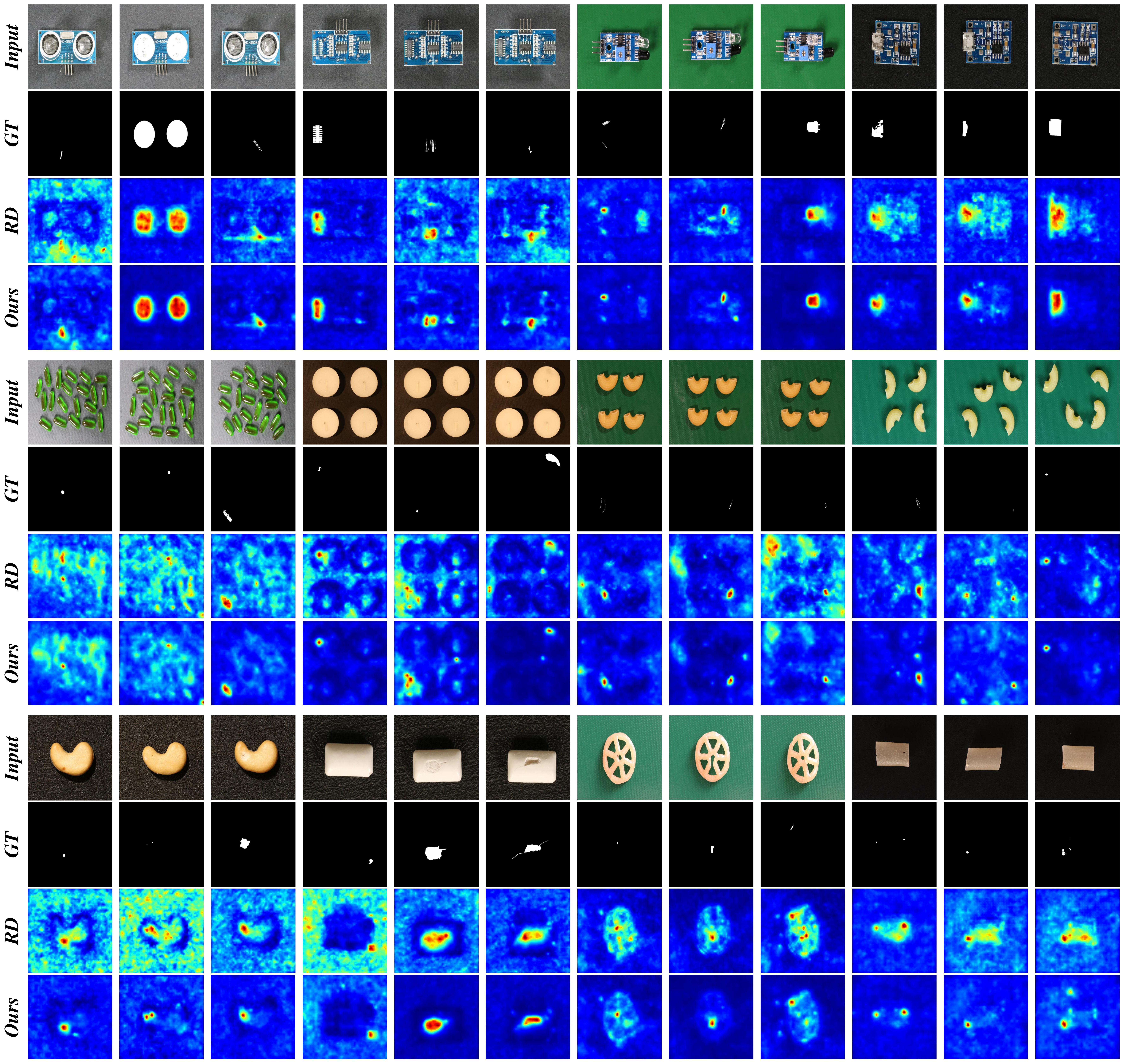} 
\caption{Visualization of anomaly localization on VisA.
}
\label{fig_visa}
\end{figure*}

On Page 1 of the main text, we provide visualizations of localization results on MVTec AD. In this section, we show qualitative results of anomaly localization on three other datasets. Figures \ref{fig_mpdd}, \ref{fig_btad}, and \ref{fig_visa} respectively show the visual results on MPDD, BTAD, and VisA datasets. It is evident that, compared to RD and RD++ \citesupp{tien2023revisiting}, our proposed method achieves more accurate anomaly localization while also reducing the likelihood of false positives in normal regions.

\bibliographystylesupp{aaai25}
\bibliographysupp{ref}

\end{document}